%% file: colm2024_conference.tex
\documentclass{article} %
\usepackage{colm2024_conference}

\usepackage{booktabs}
\usepackage{graphicx}
\usepackage{enumitem}
\usepackage{wrapfig}
\usepackage{algorithm}
\usepackage{algpseudocode}

\usepackage{CJKutf8}
\usepackage{microtype}
\usepackage{amsmath}
\usepackage{amsfonts}
\usepackage{colortbl}
\usepackage[utf8]{inputenc}
\usepackage[T1]{fontenc}
\definecolor{lightgray}{rgb}{0.9,0.9,0.9}
\usepackage{caption}
\usepackage{subcaption}
\usepackage{xcolor}
\usepackage{setspace}
\usepackage{url}
\usepackage{multirow}
\usepackage{colortbl}
\usepackage{tabularx}
\usepackage{xltabular}
\usepackage{blindtext}
\usepackage{pgfplots}
\pgfplotsset{compat=1.18} 
\usepackage{tikz}
\usetikzlibrary{er,positioning,bayesnet}
\usepackage{makecell}
\usepackage{tipa}
\usepackage{siunitx}
\usepackage{nicefrac}
\usepackage{tocloft}
\usepackage{listings}
\usepackage[raster,skins]{tcolorbox} %
\usepackage{xltabular}
\usepackage{adjustbox}
\usepackage{xurl}
\usepackage{multirow}
\usepackage{threeparttable}
\usepackage{xcolor}
\input{math_commands.tex}

\title{Qwen3-Omni Technical Report}

\author{
\bf Qwen Team
}

\newcommand{\tabincell}[2]{\begin{tabular}{@{}#1@{}}#2\end{tabular}}

\newcommand{\method}{Qwen3-Omni\xspace}

\begin{document}

\maketitle

\begin{abstract}
We present Qwen3-Omni, a single multimodal model that for the first time maintains state-of-the-art performance across text, image, audio, and video without any degradation relative to single-modal counterparts. Qwen3-Omni matches the performance of same-sized single-modal models within the Qwen series and excels particularly on audio tasks. Across 36 audio and audio-visual benchmarks, Qwen3-Omni achieves open-source state-of-the-art (SOTA) on 32 benchmarks and overall SOTA on 22, outperforming strong closed-source models such as Gemini-2.5-Pro, Seed-ASR, and GPT-4o-Transcribe.
Qwen3-Omni adopts a Thinker–Talker Mixture-of-Experts (MoE) architecture that unifies perception and generation across text, images, audio, and video, yielding fluent text and natural real-time speech. It supports text interaction in 119 languages, speech understanding in 19 languages and speech generation in 10 languages. The system can process audio recordings up to 40 minutes per instance for ASR and spoken-language understanding, enabling high-quality audio and audiovisual experiences across locales. It demonstrates strong instruction following and allows fine-grained customization of conversational tone and persona via user-defined system prompts.
To reduce first-packet latency in streaming synthesis, the Talker autoregressively predicts discrete speech codecs using a multi-codebook scheme. Leveraging the representational capacity of these codebooks, we replace computationally intensive block-wise diffusion with a lightweight causal ConvNet, enabling streaming from the first codec frame. In cold-start settings (no prior context), Qwen3-Omni achieves a theoretical end-to-end first-packet latency of 234 ms. 
To further strengthen multimodal reasoning, we introduce a Thinking model that explicitly reasons over inputs from any modality. Since the research community currently lacks a general-purpose audio captioning model, we fine-tuned Qwen3-Omni-30B-A3B to obtain Qwen3-Omni-30B-A3B-Captioner, which produces detailed, low-hallucination captions for arbitrary audio inputs. Qwen3-Omni-30B-A3B, Qwen3-Omni-30B-A3B-Thinking and Qwen3-Omni-30B-A3B-Captioner are publicly released under the Apache 2.0 license.

\end{abstract}

%Under industrial-scale, high-concurrency conditions (e.g., on the Qwen-Chat platform\footnote{\url{https://chat.qwen.ai/}}), the system delivers audio and audiovisual first-packet latencies of \jin{XX} seconds. 

% \vfill

% \begin{figure}[hbp]
%     \centering
%     \includegraphics[width=0.92\textwidth]{figures/main.pdf}
%     \caption{In the iterative development of the Qwen series, data scaling has played a crucial role. Qwen~2.5, which leverages 18 trillion tokens for pre-training, has demonstrated the most advanced capabilities within the Qwen series, especially in terms of domain expertise, underscoring the importance of scale together with mixture in enhancing the model's capabilities.}
%     \label{fig:intro}
% \end{figure}

% \vfill

% \newpage

% \begin{figure}[tbh]
%     \centering
%     \includegraphics[width=0.8\textwidth]{figures/qwen_omni.png}
%     \caption{\method~is a unified end-to-end model capable of processing multiple modalities, such as text, audio, image and video, and generating real-time text or speech response. Based on these features, \method~supports a wide range of tasks, including but not limited to voice dialogue, video dialogue, and video reasoning.}
%     \label{fig:ovewview_arch}
% \end{figure}

\begin{figure}[tbh]
    \centering
    \includegraphics[width=0.95\textwidth]{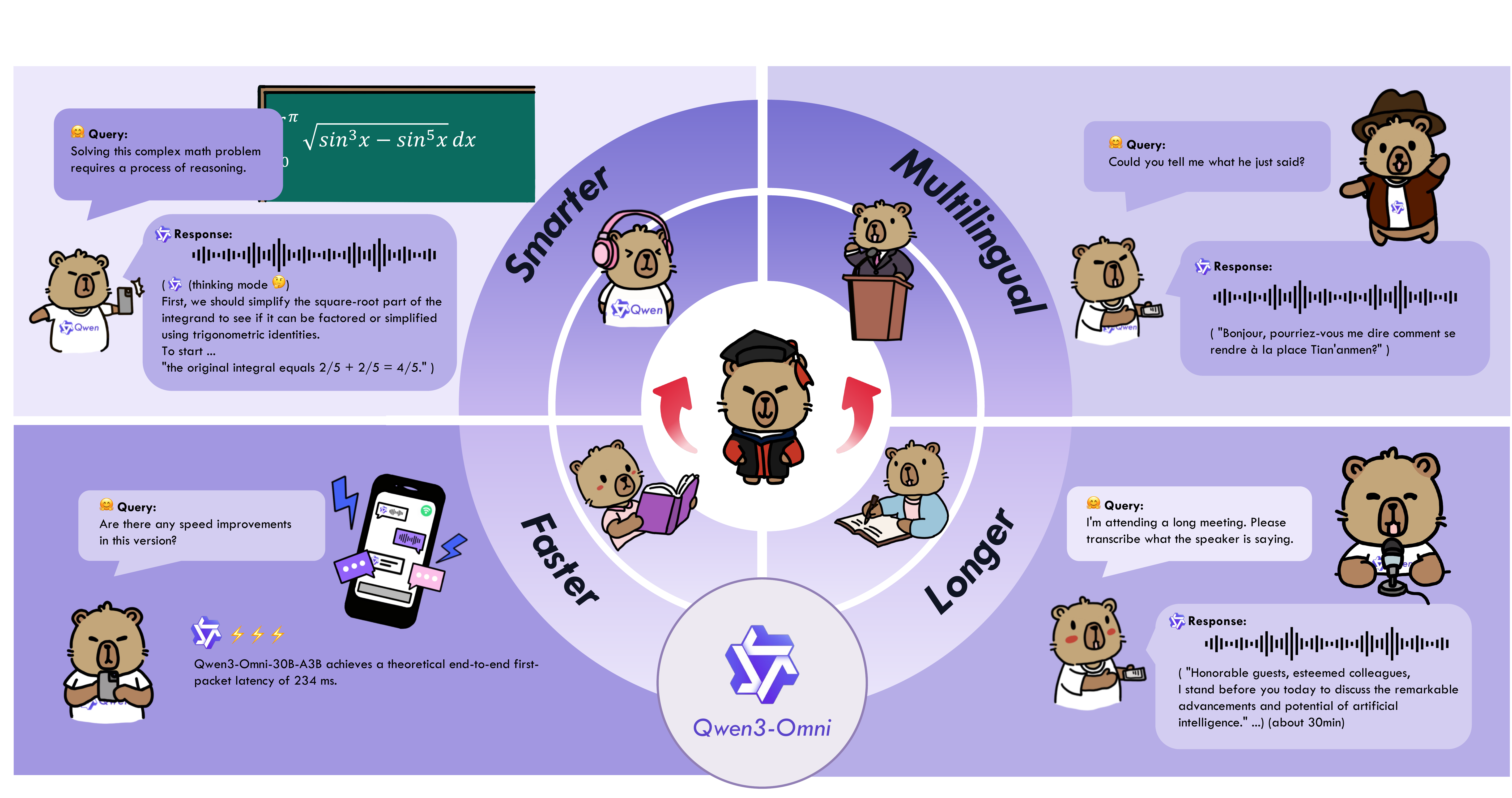}
    \caption{\method~is a unified end-to-end model capable of processing multiple modalities, such as text, audio, image and video, and generating real-time text or speech response. Based on these features, \method~supports a wide range of tasks, including but not limited to voice dialogue, video dialogue, and video reasoning.}
    \label{fig:intro}
\end{figure}

\input{content/intro.tex}

\input{content/arch.tex}
\input{content/pretraining}

\input{content/posttraining}

\input{content/experiments.tex}
\input{content/omniexp.tex}

\input{content/conclusion.tex}
\input{content/authors.tex}

\clearpage
\input{content/appendix.tex}

\clearpage
\bibliography{biblio}
\bibliographystyle{colm2024_conference}

\end{document}

%% file: math_commands.tex
\usepackage{amsmath,amsfonts,bm}

\def\eqref#1{equation~\ref{#1}}
\def\1{\bm{1}}

\DeclareMathAlphabet{\mathsfit}{\encodingdefault}{\sfdefault}{m}{sl}
\SetMathAlphabet{\mathsfit}{bold}{\encodingdefault}{\sfdefault}{bx}{n}

%% file: content/intro.tex
\section{Introduction}
\label{sec:intro}

Humans perceive visual and auditory inputs in parallel, cognitively process these signals, and emit responses through textual expression, vocalization, and tool-mediated or bodily actions, facilitating information exchange with other organisms and demonstrating intelligence. Building on the rapid advances in the understanding and reasoning capabilities of unimodal large models~\citep{gpt3,gpt4,gemini,claude,claude2,claude3,qwen,qwen2,qwen3,llama2,llama3,blip2,llava,minigpt-4,qwenvl,qwen2.5vl, qwenaudio, qwen2-audio}, natively multimodal systems have drawn substantial attention~\citep{gpt4o,gemini2.5,qwen2.5omni}. Human learning typically progresses through the coordinated use of multiple modalities, where complementary specialization and cross-modal synergy improve learning efficiency. However, contemporary LLM-centric multimodal models often exhibit modality trade-offs, with gains in one modality accompanied by degradation in others.

In this report, we take a step toward resolving this limitation by exploring integrated multimodal training within the prevailing LLM-based paradigm. We demonstrate that joint multimodal training can achieve parity across all modalities—i.e., no modality-specific performance degradation—while markedly enhancing cross-modal capabilities such as video understanding. A key ingredient is mixing unimodal and cross-modal data during the early stage of text pretraining. As evidenced by Qwen3-Omni-30B-A3B-Base, its text and vision performance is on par with same-sized single-modal text and vision base models across extensive benchmarks, while simultaneously exhibiting strong audio competence, audiovisual understanding, cross-modal ``thinking'', and real-time audiovisual interaction. The development of non-degrading multimodal systems is an achievable objective. Such systems are characterized by two key properties: first, their ability to match the performance of specialized unimodal models in their respective tasks, and second, their capacity to facilitate novel cross-modal reasoning and interaction. These latter capabilities represent a significant advantage, as they are not present in traditional unimodal approaches.

% Although multiple routes may lead to non-degrading multimodality, 
% we want to make clear to researchers and practitioners that non-degrading multimodality is attainable: native multimodal models can match modality-specific performance across all modalities and, moreover, provide cross-modal reasoning and interaction capacities unavailable to unimodal counterparts.

Qwen3-Omni builds on the Thinker–Talker architecture introduced in Qwen2.5-Omni~\citep{qwen2.5omni} and introduces \textbf{five key upgrades}: (1) both the Thinker and Talker are upgraded to Mixture-of-Experts (MoE) designs; (2) we replace Whisper audio encoder with our AuT (Audio Transformer) encoder, trained from scratch on 20 million hours of supervised audio, yielding stronger general-purpose audio representations. AuT employs block-wise window attention to enable real-time prefill caching; (3) on the speech generation side, we adopt a multi-codebook representation,
% , named \textit{Qwen3-TTS-Tokenizer-v2}, 
whose increased capacity supports faithful modeling of diverse voices, paralinguistic cues, and acoustic phenomena; (4) the Talker shifts from single-track to multi-track codec modeling, autoregressively predicting multiple codebook layers via MTP modules, while the waveform stage (Code2Wav) replaces block-wise DiT with a lightweight convolutional network (ConvNet); and (5) the input and output audio code rates are reduced to 12.5 Hz, with the output codec enabling single-frame, immediate speech synthesis. Taken together, these changes enable low-latency speech interaction under high concurrency in industrial-scale deployments.

Compared with Qwen2.5-Omni, Qwen3-Omni introduces \textbf{four major improvements}: (1) support for audio understanding on inputs exceeding 40 minutes; (2) expanded language coverage to 119 written languages, 19 and 10 spoken languages for understanding and generation respectively ; (3) a Thinking model enabling full-modality reasoning, including audio–video and audio-only scenarios; and (4) improved streaming performance with end-to-end latency as low as 234 ms.
% Discuss the differences between Qwen3-omni and Qwen2.5-Omni
% 1 长音频和长视频
% 2 帧率
% 3 多语言文本，语音理解
% 4 数据量
% 5 生成
% 5.1 多语言语音生成
% 5.2 极致流式 - 多码本 with sub-Talker
% 5.3 跨语种迁移能力
% 6 全模态不降智
% 7 交互效率
% 8 Thinking
% Compared to its predecessor, Qwen2.5-Omni, the Qwen3-Omni model incorporates four principal advancements:

% \begin{itemize}
%     \item \textbf{Extended Audio Processing:} The model's capacity for audio comprehension has been enhanced to process long-form inputs with durations exceeding 40 minutes.
    
%     \item \textbf{Expanded Multilingualism:} Language support has been significantly broadened to encompass 119 written languages. Furthermore, spoken language capabilities now include 19 languages for understanding and 10 for generation.
    
%     \item \textbf{Advanced Multimodal Reasoning:} A specialized 'Thinking' model has been integrated to facilitate full-modality reasoning across complex scenarios, including those involving combined audio-video and audio-only data.
    
%     \item \textbf{Improved Streaming Latency:} Streaming performance has been optimized, achieving a reduced end-to-end latency with a minimum response time of 234 milliseconds (ms).
% \end{itemize}

Critically, Qwen3-Omni maintains state-of-the-art performance on text and visual modalities without degradation relative to same-size single-model Qwen counterparts. Across 36 audio and audio-visual benchmarks, it achieves open-source SOTA on 32 and sets the SOTA on 22, outperforming strong closed-source systems such as Gemini 2.5 Pro, Seed-ASR, and GPT-4o-Transcribe.

The remainder of this paper is organized as follows. Section~2 presents the algorithms and architecture of Qwen3-Omni. Sections~3 and~4 describe the pretraining and post-training datasets and pipelines, respectively. Section~5 reports the experimental results. Section~6 compares Qwen3-Omni with recent Qwen models of comparable parameter scales, demonstrating multimodal performance without modality-induced degradation. 
%Section~7 presents an application case in which we fine-tune Qwen3-Omni to obtain Qwen3-Omni-Captioner, a model that generates highly detailed captions for general audio.

%% file: content/arch.tex
\section{Architecture}
\begin{figure}[tbh]
    \centering
    \includegraphics[width=0.8\textwidth]{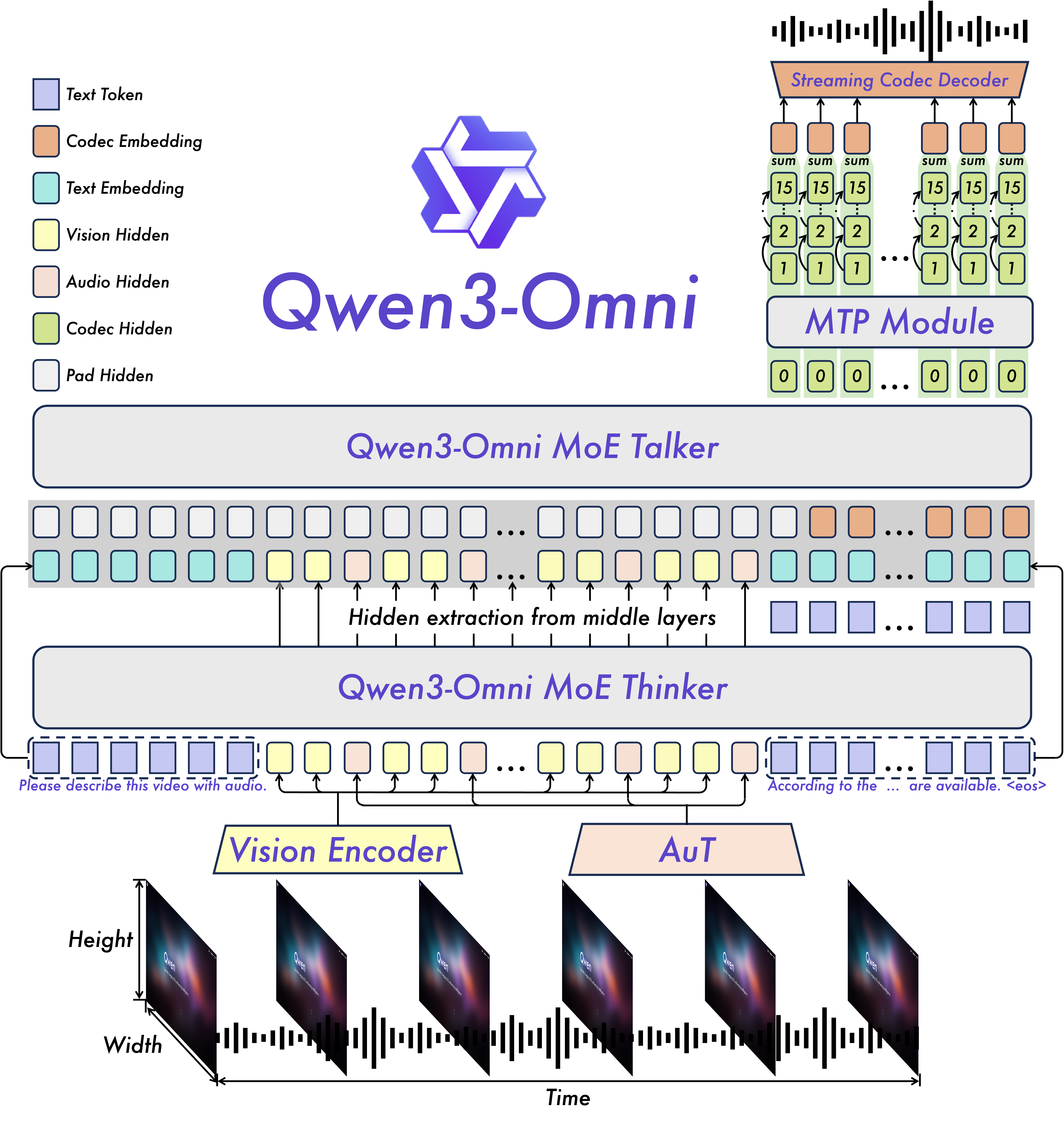}
    \caption{The overview of \method. \method adopts the Thinker-Talker architecture. Thinker is tasked with text generation while Talker focuses on generating streaming speech tokens by receives high-level representations directly from Thinker. To achieve ultra–low-latency streaming, Talker autoregressively predicts a multi-codebook sequence. At each decoding step, an MTP module outputs the residual codebooks for the current frame, after which the Code2Wav renderer incrementally synthesizes the corresponding waveform, enabling frame-by-frame streaming generation.}
    \label{fig:ovewview_arch}
\end{figure}

\subsection{Overview}
As shown in Figure~\ref{fig:ovewview_arch}, \method employs Thinker-Talker architecture~\citep{qwen2.5omni}. Compared with Qwen2.5-Omni, \method introduces the following changes for greater scalability and control:
\begin{itemize}
    \item Both the Thinker and Talker adopt Mixture-of-Experts (MoE) architectures to support high concurrency and fast inference.
    \item Talker no longer consumes the Thinker’s high-level text representations and conditions only on audio and visual multimodal features. This design is motivated by: (i) for textual content, discrete tokens and embeddings are effectively information-equivalent; and (ii) multimodal conditioning is necessary for audio–video–coordinated speech generation such as preserving prosody/timbre in speech translation. Moreover, this decoupling allows external modules (e.g., RAG, function calling, safety filters) to intervene on the Thinker’s textual output and, if desired, supply text to the Talker via controlled preprocessing for streaming synthesis.
    \item Since textual representations are decoupled, the Thinker and Talker can use distinct system prompts, independently controlling the Thinker’s response style and the Talker’s audio style.
    \item The Talker adopts a multi-codebook autoregressive scheme: Talker generates one codec frame per step, while the MTP module produces the remaining residual codebooks.
    \item The Code2Wav is implemented as a lightweight causal ConvNet,  simplifying the final stage of audio synthesis.
    % \item The Code2Wav is simplified to a lightweight ConvNet.
\end{itemize}
%1) Both the Thinker and Talker adopt Mixture-of-Experts (MoE) architectures to support high concurrency and fast inference. 2) The Talker no longer consumes the Thinker’s high-level text representations and conditions only on audio and visual multimodal features. This design is motivated by: (i) for textual content, discrete tokens and embeddings are effectively information-equivalent, whereas multimodal features are more comprehensive; and (ii) multimodal conditioning is necessary for audio–video–coordinated speech generation and for preserving prosody/timbre in speech translation. Moreover, this decoupling allows external modules (e.g., RAG, function calling, safety filters) to intervene on the Thinker’s textual output and, if desired, supply text to the Talker via controlled preprocessing for streaming synthesis. 3) Since textual representations are decoupled, the Thinker and Talker can use distinct system prompts, independently controlling the Thinker’s response style and the Talker’s audio style. 4) The Talker adopts a multi-codebook autoregressive scheme: Talker generates one codec frame per step, while sub-talkers produce the remaining residual codebooks. 5) The Code2Wav renderer is simplified to a lightweight ConvNet.

During training and inference, the Talker directly ingests high-dimensional multimodal features from the Thinker and shares access to the full conversational history. As a result, the system operates as a cohesive single model, enabling end-to-end training and unified inference.

In the following sections, we first introduce with our newly proposed AuT encoder, including its training methodology. Then, describe how Thinker processes various inputs. We then detail Talker’s multi-codebook streaming speech generation. Finally, we highlight a series of improvements on both the understanding and generation modules aimed at achieving ultra–low-latency, end-to-end streaming audio inference.

\subsection{Audio Transformer~(AuT)}

\begin{figure}[tbh]
    \centering
    \includegraphics[width=0.6\textwidth]{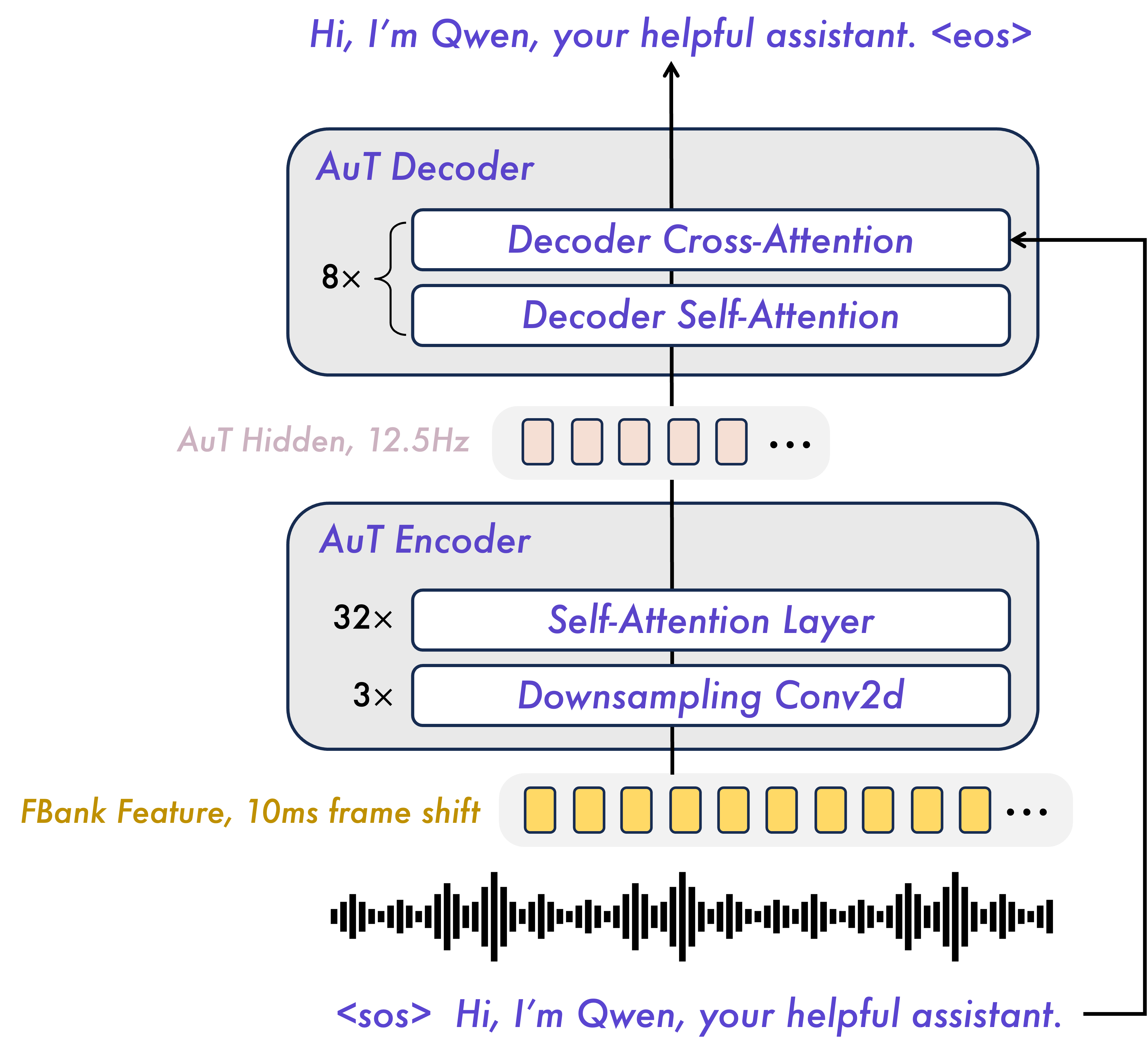}
    \caption{The overview of AuT. AuT is an attention-encoder-decoder based auto-regressive model, which is trained from scratch on 20 million hours of supervised audio. \method employs the AuT encoder as the audio encoder to obtain general purpose audio representations at a token rate of 12.5Hz.}
    \label{fig:aut}
\end{figure}

Audio Transformer (AuT) is an attention-encoder-decoder model, as is shown in Figure~\ref{fig:aut}, trained from scratch on 20 million hours of supervised audio data. During training, the filter bank features of the audio are downsampled 8 times using Conv2D blocks before the attention layers, reducing the token rate to 12.5 Hz. To learn stronger and more general-purpose audio representations, AuT is trained on large-scale audio datasets with both speech recognition and audio understanding tasks. Specifically, the training data includes 80\% Chinese and English pseudo-labeled ASR data, 10\% ASR data from other languages, and 10\% audio understanding data. To balance the efficiency of real-time prefill caching with the performance for offline audio tasks, AuT utilizes flash attention with dynamic attention window sizes, covering attention query patterns ranging from 1 to 8 seconds. In \method, we employ the AuT encoder as the audio encoder, which contains approximately 0.6B parameters.

\subsection{Perceivation}
% \begin{figure}[tbh]
%     \centering
%     \includegraphics[width=1.0\textwidth]{figures/ATMRoPE.png}
%     \caption{An illustration of Time-aligned Multimodal RoPE~(TMRoPE).}
%     \label{fig:ATMRoPE}
% \end{figure}
\paragraph{Text, Audio, Image and Video (w/o Audio).} 

Thinker converts text, audio, image, and video (without audio) into a series of  representations for input. For text inputs, we use Qwen’s tokenizer~\citep{qwen3}, which applies byte-level byte-pair encoding with a vocabulary of 151,643 regular tokens. For audio inputs and audio extracted from video, we resample to 16 kHz and convert the raw waveform into a 128 channel mel-spectrogram with a 25 ms window and a 10 ms hop. We adopt AuT encoder as our audio encoder, which is trained from scratch on 20 millions hours of audio data, and each frame of the audio representation corresponds to approximately an 80 ms segment of the original audio signal. Furthermore, we employ the vision encoder from Qwen3-VL, initialized from SigLIP2-So400m~\citep{siglip2} with approximately 543 million parameters, enabling handling of both image and video inputs. The vision encoder is trained on a mixture of image and video data, ensuring strong image understanding and video comprehension. To preserve video information as completely as possible while aligning with the audio sampling rate, we sample video frames at a dynamic frame rate.

\paragraph{Video and Multimodal Position Embedding~(TM-RoPE)}  

Drawing inspiration from Qwen2.5-Omni, we employs a Time-aligned Multimodal Rotary Position Embedding (TM-RoPE), which extends the Multimodal Rotary Position Embedding (M-RoPE)~\citep{qwenvl} by incorporating absolute temporal information. TM-RoPE factorizes the conventional rotary position embedding into three distinct dimensions: temporal, height, and width. In the original M-RoPE formulation, temporal dependencies are modeled using the initial 16 rotary angles, which correspond to higher frequencies and exhibit stronger oscillatory patterns. While this design is effective for capturing fine-grained local temporal variations, it can impede the model's ability to extrapolate over extended sequences. To address this limitation, we introduce a modified allocation of rotary angles. Specifically, the temporal, height, and width dimensions are interleaved and assigned 24, 20, and 20 rotary angles, respectively. This redistribution fosters a more balanced representation of both local semantics and long-range dependencies, thereby enhancing the model's overall performance. The application of TM-RoPE is tailored to the specific modality of the input data. For text inputs, the three components share identical position identifiers, rendering TM-RoPE functionally equivalent to a one-dimensional RoPE~\citep{rope}. Similarly, audio inputs utilize shared position IDs but are further augmented with absolute temporal encodings, where each temporal ID corresponds to a duration of 80 ms. For image data, a constant temporal ID is assigned to all visual tokens, while their distinct row and column positions determine the height and width IDs.

In the context of multimodal audiovisual streams, the audio component is encoded with a temporal ID for every 80 ms. The video is treated as a sequence of frames with monotonically increasing temporal IDs that are dynamically adjusted based on their actual timestamps to ensure a consistent temporal resolution of 80 ms per ID. The height and width IDs for video frames are assigned in the same manner as for still images. To prevent positional conflicts when processing multiple modalities, the position numbering is made contiguous, with each subsequent modality commencing from one plus the maximum position ID of the preceding modality. This refined approach to positional encoding enables the model to effectively integrate and jointly model information from diverse modalities. In a departure from Qwen2.5-Omni, which segments audiovisual representations into fixed 2-second chunks, \method directly aligns these representations using their temporal IDs, which are explicitly anchored to absolute time. This design choice affords the model the flexibility to support streaming inputs of arbitrary duration.

\subsection{Speech Generation}
% In multi-turn dialogue, Talker directly inherits the Thinker’s context, including textual tokens and multimodal high-level representations, and synthesizes speech conditioned on the Thinker’s streamed textual output for the current turn. As with text generation, intelligent speech synthesis cannot rely solely on the current query. Long-context representations enable the model to adapt fine-grained acoustic attributes—such as prosody, loudness, and emotion—based on the broader discourse context.

% Unlike~\citet{qwen2.5omni}, Talker operates directly on RVQ tokens produced by \textit{Qwen3-TTS-Tokenizer-v2}. Talker backbone ingests the aggregated codebook features of the current frame and uses a linear head to predict the zeroth codebook. Then, a multi-token prediction (MTP) module then generates the residual codebooks. This design allows Talker to model the full spectrum of fine-grained acoustic detail, thereby enhancing expressivity. Since Talker learns an end-to-end representation over the complete set of codebooks, Code2Wav requires only a causal ConvNet to reconstruct the waveform, achieving substantially lower inference latency and FLOPs while delivering higher fidelity than DiT-based decoders.

For speech synthesis in multi-turn dialogues, our Talker module is conditioned on a rich context inherited from a "Thinker" component, comprising historical textual tokens, multimodal representations, and the current turn's streamed text. This reliance on long-context information is critical, as high-fidelity speech synthesis must adapt acoustic attributes like prosody, loudness, and emotion to the ongoing discourse, a principle well-established in context-aware generative models.

Architecturally, our approach departs from~\citet{qwen2.5omni} by operating directly on RVQ tokens. The Talker employs a hierarchical prediction scheme: the backbone ingests the aggregated codebook features of the current frame and uses a linear head to predict the zeroth codebook, after which a multi-token prediction (MTP) module generates all residual codebooks. This strategy enables the model to learn a complete representation of acoustic details, enhancing vocal expressivity. Consequently, waveform reconstruction is simplified to a lightweight causal ConvNet (Code2Wav), which significantly reduces inference latency and computational cost (FLOPs) while achieving superior audio fidelity compared to more complex DiT-based vocoders.

\subsection{Designs for Streaming and Concurrency}
In streaming audiovisual interaction scenarios, the first-packet latency is a critical factor affecting user experience, and the model's concurrency capability is key to reducing service costs and improving response speed. This section discusses how \method enhances concurrency and reduces first-packet latency through algorithmic and architectural optimizations.

\begin{table}[htbp]
\centering
\caption{\textbf{The architectural design of Qwen3-Omni-30B-A3B and the end-to-end first-packet latency for Audio/Video~(ms).}}
\vspace{-1mm}
\begin{tabular}{lccc}
\toprule
\textbf{Module}         & \textbf{Architecture}      & \textbf{Params}    & \textbf{Streaming} \\\midrule
Audio Encoder  & AuT               & 650M      & $\checkmark$        \\
Vision Encoder & SigLIP2-So400M    & 540M     & -        \\
Thinker        & MoE Transformer   & 30B-A3B  & $\checkmark$        \\
Talker         & MoE Transformer   & 3B-A0.3B & $\checkmark$        \\
MTP            & Dense Transformer & 80M     & $\checkmark$        \\
Code2wav       & ConvNet           & 200M     & $\checkmark$        \\\midrule
\multicolumn{4}{c}{End-to-End First-Packet Latency: \textbf{234/547ms}} \\
\bottomrule
\end{tabular}
\end{table}

\paragraph{Chunked Prefilling and MoE Architecture.} In \method, we retain the chunked-prefilling mechanism as implemented in Qwen2.5-Omni, whose audio and vision encoders are capable of outputting chunks along the temporal dimension. During real-time interaction, Thinker and Talker modules perform asynchronous prefilling: when Thinker completes prefilling the current chunk, its output high-level representations are immediately used to prefill the Talker’s current chunk asynchronously, while Thinker prefills its next chunk. This approach significantly reduces the Time-To-First-Token (TTFT) for both the Thinker and the Talker. Architecturally, both Thinker and the Talker in \method adopt the MoE design, which is highly effective for improving service throughput. Compared to dense models, the MoE architecture significantly decreases IO consumption arising from KV cache during processing of long sequences, thereby increasing tokens per second (TPS) during generation and enhancing concurrency.

\paragraph{Streaming Multi-Codebook Codec Generation.} To minimize the user's waiting time for receiving the first generated packet, we propose a \textit{left context only multi-codebook generation} mechanism. As shown in Figure~\ref{fig:ovewview_arch}, once Talker generates the first token, the MTP module predicts the remaining tokens for the current frame. These tokens are then decoded into waveform by a streaming multi-codebook codec decoder that only attends to the left context. Unlike Qwen2.5-Omni that requires waiting for sufficient block-context from the Talker before synthesis, \method can output the waveform immediately after the Talker generates each token, significantly reducing first-packet latency. 

\paragraph{Lightweight MTP module and ConvNet.} Both the MTP module and codec decoder are lightweight modules, which have low computational FLOPs and support batched inference, making them well-suited for high-concurrency scenarios. The MTP Module is an ultra-lightweight fixed-step autoregressive dense transformer, with low memory bandwidth requirements on inference hardware, thereby naturally enabling efficient batch processing of high throughput requests. Its fixed-step autoregressive inference mechanism allows it to effectively leverage a fixed KV cache memory space for acceleration, achieving low inference latency. Meanwhile, the ConvNet-based codec decoder also achieves high throughput with low latency because its convolutional architecture enjoys extensive hardware acceleration support across diverse inference platforms, and it enables efficient batched inference.

\begin{table}[htbp]
\centering
\caption{\textbf{Theoretical First-Packet Latency of \method wit Different Concurrency.}}
\label{tab:inference-lantency}
\small
\setlength{\tabcolsep}{2.6pt}
\begin{tabular}{@{}lccc@{}}
\toprule
 & \multicolumn{3}{c}{\textbf{\method-30B-A3B}} \\
\cmidrule(lr){2-4}
 & \textbf{1 Concurrency} & \textbf{4 Concurrency} & \textbf{6 Concurrency} \\
\midrule
Thinker-Talker Tail Packet Preprocessing Latency & 72/160ms & 94/180ms & 100/200ms \\
Thinker Time-to-First-Token (TTPT) & 88/160ms & 468/866ms & 673/1330ms \\
Talker Time-to-First-Token (TTPT) & 57/210ms & 145/450ms & 376/734ms \\
MTP Module Time Cost Per Token & 14ms & 16ms & 18ms\\
Codec Decoder Time Cost Per Code & 3ms & 5ms & 5ms \\
\midrule
\textbf{Overral Latency~(Audio/Video)} & \textbf{234/547ms} & \textbf{728/1517ms} & \textbf{1172/2284ms} \\
\midrule
Thinker Token Generation Rate (TPS) & 75 tokens/s & 63 tokens/s & 53 tokens/s \\
Talker Token Generation Rate (TPS) & 140 tokens/s & 125 tokens/s & 110 tokens/s \\
\midrule
\textbf{Generation RTF(Real Time Factor)} & \textbf{0.47} & \textbf{0.56} & \textbf{0.66} \\
\bottomrule
\end{tabular}
\end{table}

Table~\ref{tab:inference-lantency} presents the theoretical first-packet latency for \method under typical computational resources across varying concurrency scenarios. Experiments are conducted on the vLLM framework~\citep{vllm} to process concurrent audiovisual streams, with optimizations applied via \textit{torch.compile} and CUDA Graph acceleration to the MTP Module and codec decoder. Several factors influence the total first-packet latency. First, the model sizes of Thinker and Talker impact their tail packet preprocessing latency (multi-modal data preprocessing and inference for Audio and Vision Encoder) and Time-To-First-Token (TTPT). Second, the architectures and sizes of the MTP Module and Codec Decoder affect their inference latency. Due to the sequential dependency between these components, the total first-packet latency represents the sum of these individual latencies. As shown in the results, the MoE architecture of Thinker and Talker ensures that their prefill latency and TTPT remain largely unaffected under high concurrency. Meanwhile, the lightweight design of the MTP Module and Codec Decoder minimizes their computational overhead, resulting in a lower impact on first-packet latency. Furthermore, after the initial packet is output and the model starts streaming audio synthesis, the 12.5Hz token rate Talker requires only one token to synthesize 80ms audio. Consequently, the Generation Real Time Factor (RTF) is calculated by dividing the sum of:  (1) the time taken by Thinker and Talker to generate one token; and (2) the processing time per token for the MTP Module and Codec Decoder by 80ms. As demonstrated, the RTF consistently remains below 1 across varying concurrency levels, ensuring that users receive continuously streaming audio responses.

% \paragraph{Large Thinker and Small Talker.} To reduce the computational cost and latency, we use a small Talker, xxx M.

% For tokenization, we utilize Qwen's tokenizer~\citep{qwen}, which implements byte-level byte-pair encoding (BBPE,~\citealp{gpt3,wang2020neural,sennirch2016neural}) with a vocabulary of 151,643 regular tokens. We have expanded the set of control tokens from 3 to 22 compared to previous Qwen versions, adding two new tokens for tool functionality and allocating the remainder for other model capabilities. This expansion establishes a unified vocabulary across all Qwen2.5 models, enhancing consistency and reducing potential compatibility issues.

% \begin{table}[tbp]
% \caption{Model architecture and license of Qwen2.5 open-weight models.}
% \small
% \centering
% \begin{tabular}{@{}lcccccccc@{}} 
% \toprule
% Models  & Layers & Heads (Q / KV) & Tie Embedding & Context / Generation Length  & License \\
% \midrule
% 0.5B  & 24 & 14 / 2 & Yes & 32K / 8K & Apache 2.0 \\
% 1.5B  & 28 & 12 / 2 & Yes & 32K / 8K & Apache 2.0 \\
% 3B  & 36 & 16 / 2 & Yes & 32K / 8K & Qwen Research \\
% 7B  & 28 & 28 / 4 & No & 128K / 8K & Apache 2.0 \\
% 14B & 48 & 40 / 8 & No & 128K / 8K & Apache 2.0 \\
% 32B & 64 & 40 / 8 & No & 128K / 8K & Apache 2.0 \\
% 72B  & 80 & 64 / 8 & No & 128K / 8K & Qwen \\
% \bottomrule
% \end{tabular}
% \end{table}

%% file: content/pretraining.tex
\section{Pretraining}
\begin{table}[htbp]
\centering
\caption{\textbf{Languages and dialects support of Qwen3-Omni-30B-A3B.}}
\vspace{-1mm}
\begin{tabular}{lccc}
\toprule
Modality         & \# Langs      & Languages   
\\\midrule
Text  & 119         & See Qwen3 for the full list.        \\
Speech Input & 19    & ar, de, en, es, fr, id, it, ja, ko, ms, nl, pt, ru, th, tr, ur, vi, yue, zh           \\
Speech Output       & 10   & de, en, es, fr, it, ja, ko, pt, ru, zh          \\
\bottomrule
\end{tabular}
\end{table}
\label{table:languages}
\method is pre-trained on a diverse dataset that encompasses multiple languages and dialects as shown in Table~\ref{table:languages} and modalities, including image-text, video-text, audio-text, video-audio, video-audio-text, and pure text corpora. Unlike Qwen2.5-Omni, which uses a single prompt for each task, we employ a wider range of natural language prompts to enhance both the generalization ability and instruction-following capabilities. To achieve robust performance across all modalities, our training strategy incorporates both unimodal and cross-modal data from the early pretraining stage.
% To ensure that joint multimodal training achieves state-of-the-art performance across all modalities, we incorporate both unimodal and cross-modal data during the early stages of text pretraining. 

% \method consists of three training stages. 
The pre-training of \method is structured into three distinct stages. In the first stage, we lock the LLM parameters and focus on training the vision and audio encoders, utilizing a vast corpus of audio-text and image-text pairs to enhance semantic understanding within the LLM. In the second stage, we unfreeze all parameters and train with a wider range of multimodal data for more comprehensive learning. In the final stage, we use data with a sequence length of 32,768 to enhance the model's ability to understand complex long-sequence data:

\begin{enumerate}[label=(\arabic*)]
    \item \textbf{Encoder Alignment Stage (S1)}: During the initial pretraining phase, the LLM component of \method is initialized with parameters from Qwen3~\citep{qwen3}, while the vision encoder is adopted from Qwen3-VL, and the audio encoder is initialized with AuT. The two encoders are trained separately on the fixed LLM, with both initially focusing on training their respective adapters before training the encoders. We abandon the stage used in~\cite{qwen2.5vl, qwen2.5omni} where the encoder and adapter are trained jointly while keeping the LLM frozen, because this approach may cause the encoder to compensate for the limitations of the frozen LLM, which can lead to degraded perception capabilities.
    \item \textbf{General Stage (S2)}: The second phase of pretraining utilizes a large-scale dataset containing approximately 2 trillion tokens, with the following distribution across modalities: text (0.57 trillion), audio (0.77 trillion), image (0.82 trillion), video (0.05 trillion), and video-audio (0.05 trillion). During this stage, the introduction of more diverse multimodal data and tasks enhances the model’s understanding and interaction capabilities in auditory, visual, textual, and audiovisual information.
    \item \textbf{Long Context Stage (S3)}: In the final pre-training phase, we increased the maximum token length from 8,192 to 32,768 and also raised the proportion of long audio and long video in the training data. Experimental results indicate that these adjustments lead to significant improvements in the model's ability to understand long sequence data.
\end{enumerate}

%% file: content/posttraining.tex
\section{Post-training}\label{sec:post}
\subsection{Thinker}

The post-training phase comprises a three-stage training process for Thinker, enabling \method to possess instruction-following capabilities. The dataset, designed in the ChatML~\citep{chatml} format, includes pure text-based dialogue data, visual modality conversation data, audio modality conversation data, and mixed-modality conversation data.

In the first stage, we introduce a lightweight Supervised Fine-Tuning (SFT) to bridge the gap between pretrained representations and downstream task requirements through targeted instruction optimization. SFT deliberately diverges from the pretraining data schema while maintaining architectural consistency with the pretrained model, enabling efficient knowledge transfer and preserving the completeness of the pretrained features.

The second stage adopts the Strong-to-Weak Distillation pipeline as described in Qwen3~\citep{qwen3} to further improve model performance. This distillation process consists of two main phases:

\begin{enumerate}[label=(\arabic*)]
    \item \textbf{Off-policy Distillation}: In the initial phase, outputs generated by teacher models are combined to provide response distillation. This helps lightweight student models acquire fundamental reasoning abilities, establishing a strong foundation for subsequent on-policy training.
    \item \textbf{On-policy Distillation}: In the second phase, the student model generates the responses based on sampled prompts. These on-policy sequences are then used for fine-tuning, where the student’s predicted logits are aligned with those of a teacher model (Qwen3-32B or Qwen3-235B-A22B) by minimizing the KL divergence.
\end{enumerate}

Finally, we leverage GSPO~\citep{gspo} to comprehensively enhance the model's capabilities and stability across various modalities, including text, image, video, and audio. To provide feedback for the aforementioned modalities, we employ two different types of rewards:

\begin{itemize}
    \item \textbf{Rule-based Reward}: For verifiable multimodal tasks (e.g., mathematics, coding, instruction following), the reward signal is derived from a set of predefined rules. Well-designed rule-based rewards can assess the correctness of model outputs with high precision, preventing issues like reward hacking.

    \item \textbf{Model-based Reward}: To assess performance on multimodal tasks that lack objective, predefined evaluation metrics, we adopt an LLM-as-a-judge protocol. The role of the automated evaluator is filled by Qwen3 for general tasks, while the specialized vision-language model, Qwen2.5-VL, is used for visually-grounded tasks. To ensure a more robust and grounded assessment, the LLM evaluator is furnished with the corresponding ground-truth or reference answer for a given query, where applicable.
\end{itemize}

% For multimodal tasks where objective evaluation via predefined rules is not feasible, we employ a large language model (LLM), such as Qwen3, to assess the quality of our model's responses. For tasks with a visual component, this evaluation is specifically conducted by Qwen2.5-VL. In instances where a ground-truth or reference answer exists for a query, it is also supplied to the evaluator for reference.

\subsection{Talker}
We introduce a four-stage training process for Talker, enabling Qwen3-Omni to generate speech response simultaneously with text. All training data is structured in the ChatML format to ensure consistency with Thinker.

In the first stage, we leverage hundreds of millions of speech data with multimodal context to train Talker, establishing a monotonic mapping from multimodal representation to speech. In the second stage, we perform Continual Pretraining (CPT) with high-quality data, which alleviates hallucinations caused by noisy data in the first stage and significantly improve the quality of generated speech. Concurrently, we perform long-context training that enhances Talker’s ability to process extended and complex inputs and generate contextually appropriate speech response. In the third stage, to improve the generalization of multilingual speech generation and system stability, we construct preference pairs from diverse multilingual speech samples and optimize the model using Direct Preference Optimization (DPO)~\citep{rafailov2024direct}. Finally, we apply speaker fine-tuning on the aforementioned base model, enabling Talker to adopt specific voices while refining the naturalness, expressiveness, and controllability of its speech response.

\subsection{Captioner}
Captioning is a foundational task in multimodal understanding, integral to the training and evaluation of large multimodal models. However, the vast majority of existing research has concentrated on visual captioning, largely neglecting the audio modality. This omission is significant, as auditory perception is a crucial component of human sensory experience and interaction with the world. To address this gap and facilitate more comprehensive research in multimodal perception, we introduce the Qwen3-Omni-30B-A3B-Captioner. This model was developed by fine-tuning the Qwen3-Omni-30B-A3B on a large-scale dataset of detailed audio descriptions. The resulting system generates detailed, low-hallucination captions for arbitrary audio inputs. The \textbf{Appendix}~\ref{sec:caption-case} provides qualitative results that demonstrate our model's captioning capabilities across diverse acoustic scenarios.

%% file: content/experiments.tex
\section{Evaluation}
\label{sec:experiment}
% We conduct comprehensive evaluation of \method, including Qwen3-Omni-30B-A3B-Instruct, Qwen3-Omni-30B-A3B-Thinking, and our internal Qwen3-Omni-Flash-Instruct and Qwen3-Omni-Flash-Thinking, which is optimized both in efficacy and efficiency with new features like supporting of dialects. 
A comprehensive evaluation was performed on a suite of models, including Qwen3-Omni-30B-A3B-Instruct, Qwen3-Omni-30B-A3B-Thinking, and two in-house developed variants, designated Qwen3-Omni-Flash-Instruct and Qwen3-Omni-Flash-Thinking. These ``Flash'' models were designed to improve both computational efficiency and performance efficacy, integrating new functionalities, notably the support for various dialects.
The evaluation results are divided into two main categories: understanding~(X$\to$Text) and speech generation~(X$\to$Speech). 

\subsection{Evaluation of X$\to$Text}
In this section, we evaluate \method's ability to comprehend various multimodal inputs (text, audio, vision, and audiovisual video) and generate textual responses.

% MMLU-Pro \citep{mmlupro}, , and SuperGPQA \citep{supergpqa}
% , and LiveBench~(2024-11-25) \citep{livebench}
% LiveCodeBench~(v6, 2025.02-2025.05) \citep{livecodebench} and 
% , TAU2~\citep{barres2025tau2}
% , MMLU-ProX~\citep{mmluprox}, INCLUDE \citep{include}, 
\paragraph{Text$\to$Text} Our evaluation of \method on text $\to$ text primarily focuses on general tasks, reasoning ability, coding ability, alignment tasks, agent, and multilingual tasks. 
Specifically, we utilize MMLU-Redux \citep{mmluredux} and GPQA \citep{gpqa} for general tasks, AIME25~\citep{aime} and ZebraLogic~\citep{zebralogic} for reasoning evaluation, 
MultiPL-E \citep{multiple} for coding, IFEval \citep{ifeval}, Creative Writing V3 \citep{creative_writing} and WritingBench \citep{writingbench} for alignment tasks, 
BFCL-v3 \citep{bfcl} for agent evaluation, 
MultiIF~\citep{multiif} and PolyMath \citep{polymath} for multilingual tasks.

% The evaluation of \method for
% We use MMLU-Pro~\citep{mmlupro}, MMLU-redux~\citep{mmluredux} and Livebench0803~\citep{livebench} for general evaluation, 
% \begin{itemize}
%     \item \textbf{General Tasks.} MMLU-Pro~\citep{mmlupro}, MMLU-redux~\citep{mmluredux} and Livebench0803~\citep{livebench}.
%     \item \textbf{Mathematics \& Science Tasks.} GPQA~\citep{gpqa}, GSM8K~\citep{gsm8k} and MATH~\citep{math}.
%     \item \textbf{Coding Tasks.} HumanEval~\citep{humaneval}, MBPP~\citep{mbpp}, MultiPL-E~\citep{multiple} (Python, C++, JAVA, PHP, TypeScript, C\#, Bash, JavaScript) and LiveCodeBench 2305-2409~\citep{livecodebench}.
%     \item \textbf{Alignment Tasks.} IFEval~\citep{ifeval}.
% \end{itemize}

% , MMAR~\citep{mmar},
\paragraph{Audio$\to$Text} The evaluation can be categorized into basic audio tasks, including Automatic Speech Recognition (ASR), Speech-to-Text (S2TT), and Music Understanding, as well as advanced audio tasks, including Voice Chatting and Audio Reasoning.
For music understanding, we use RUL-MuchoMusic \citep{zang2025you} for a comprehensive evaluation of the music understanding capabilities of the model.
We utilize MMAU~\citep{sakshi2024mmaumassivemultitaskaudio} and MMSU~\citep{mmsu} for audio reasoning tasks, VoiceBench~\citep{chen2024voicebench} for voice-chatting tasks. We also employ multiple datasets including GTZAN \citep{tzanetakis2002musical}, four subsets of MTG-Jamendo (MTG, \cite{bogdanov2019mtg}), and MagnaTagATune \citep{law2009evaluation} to evaluate the model's capabilities across various music information retrieval tasks including genre identification, emotion and theme recognition, instrument recognition and music keyword annotation. We follow the evaluation set composition in MARBLE \citep{yuan2023marble} for GTZAN, MTG-Jamendo and MagnaTagATune.

%, and a self-curated speech-query benchmark
% OCRBench~\citep{liu2024ocrbenchhiddenmysteryocr}, 
% InfoVQA~\citep{Mathew2021InfographicVQA}, 
\paragraph{Vision$\to$Text}
The evaluation of the model's vision-to-text capabilities encompasses a suite of benchmarks targeting diverse and challenging tasks. To assess performance in general visual question answering, the model is evaluated on MMStar~\citep{chen2024we}, HallusionBench~\citep{hallusionbench}, and MM-MT-Bench~\citep{agrawal2024pixtral12b}. For the specialized domain of mathematical and STEM reasoning, we utilize MathVista~\citep{mathvista}, MathVision~\citep{mathvision}, MMMU~\citep{yue2023mmmu}, and MMMU-Pro~\citep{mmmupro}. The model's proficiency in document understanding is measured using the AI2D~\citep{kembhavi2016diagram} and ChartQA~\citep{masry2022chartqa} benchmarks. Furthermore, the model's numerical reasoning and counting abilities are specifically tested on CountBench~\citep{countbench}. To evaluate performance on dynamic visual data, we report results on three long video understanding benchmarks: Video-MME~\citep{fu2024video}, LVBench~\citep{lvbench}, and MLVU~\citep{mlvu}.

% Specifically, we utilize MMMU~\citep{yue2023mmmu} and MMMU-Pro~\citep{mmmupro} for college-level problems evaluation, MathVista~\citep{mathvista} and MathVision~\citep{mathvision} for math. For general visual question answering, we evaluate the performance on benchmark datasets such as MMBench-V1.1~\citep{MMBench}, MMVet~\citep{yu2024mm}, MMStar~\citep{chen2024we}, MME~\citep{fu2023mme}, MuirBench~\citep{wang2024muirbenchcomprehensivebenchmarkrobust}, CRPE~\citep{wang2024allseeingprojectv2general}, RealWorldQA~\citep{X.AI.}, 
% MMERealWorld~\citep{mme-realworld}, and MM-MT-Bench~\citep{agrawal2024pixtral12b}. 
% Additionally, we evaluate \method on various OCR benchmarks, such as AI2D~\citep{kembhavi2016diagram}, TextVQA~\citep{textvqa}, DocVQA~\citep{docvqa}, ChartQA~\citep{masry2022chartqa}, and OCRBench\_v2~\citep{fu2024ocrbenchv2improvedbenchmark}. Furthermore, we also evaluate the visual grounding capability of our model on the referring expression comprehension benchmarks~\citep{kazemzadeh-etal-2014-referitgame, mao2016generationcomprehensionunambiguousobject}, object detection in the wild~\citep{li2022groundedlanguageimagepretraining} and a self-curated point grounding benchmark. We assess our model on several representative video understanding tasks like Video-MME~\citep{fu2024video},
% MVBench~\citep{li2024mvbench}, and EgoSchema~\citep{mangalam2023egoschema}.

\paragraph{AudioVisual Video$\to$Text}
To evaluate the model's ability to process dynamic multi-modal information, we first assessed its performance on the WorldSense benchmark~\citep{worldsense}. This benchmark is designed to measure the integration of visual and auditory signals, a foundational capability for operating in complex, open-world environments. To further examine the model's higher-order cognitive functions, we then evaluated its performance on two audiovisual reasoning benchmarks: DailyOmni~\citep{dailyomni} and VideoHolmes~\citep{videoholmes}.

\subsubsection{Performance of Text$\to$Text}
We compare \method with other leading large language models (thinking or instruct). According to Table \ref{tab:text_nonthink} and \ref{tab:text_think}, 
notably, despite a smaller parameter count, \method-30B-A3B-Instruct surpasses the performance of the larger open-source model Qwen3-235B-A22B Non-Thinking and the formidable closed-source model GPT-4o-0327 across a suite of benchmarks, including GPQA, AIME25, ZebraLogic, WritingBench, and PolyMath. Concurrently, \method-30B-A3B-Thinking demonstrates performance comparable to that of Gemini-2.5-Flash-Thinking and Qwen3-235B-A22B Non-Thinking. Furthermore, \method-30B-A3B exhibits textual capabilities on par with its text-only counterparts, namely the Qwen3-30B-A3B-Instruct-2507 and Qwen3-30B-A3B-Thinking-2507.

% Furthermore, Our internal Qwen3-Omni-Flash achieves even better results.

\input{posttrain_tables/text.tex}

\subsubsection{Performance of Audio$\to$Text}
We compare \method with other leading specialist and generalist models on ASR \& S2TT, voice-chatting, audio reasoning, and music understanding benchmarks. For brevity, we defer the results of the Qwen3-Omni-Thinking model on ASR \& S2TT and music understanding to the \textbf{Appendix}~\ref{sec:audio-eval}.
\input{posttrain_tables/audio_asrs2tt.tex}

As shown in Table \ref{tab:audio_asr}, \method-Instruct achieves state-of-the-art En \& Zh ASR and lyric ASR performance on Librispeech, Wenetspeech, Fleurs, CommonVoice, Opencpop-test and MIR-1K (vocal). It also delivers better or comparable performance with other specialist or generalist models like Voxtral-Small and Gemini-2.5-Pro on Multilingual ASR and S2TT. These results show a strong performance of \method in speech recognition and speech translation.

Additionally, on VoiceBench shown in Table \ref{tab:audio_voicebench}, \method-Thinking achieves an impressive average score of 89.5, surpassing all other audio language models except Gemini-2.5-Pro (89.6). This showcases our model's strong capabilities in speech interaction. \method also demonstrates impressive performance in audio reasoning, outperforming the powerful closed-source models Gemini-2.5-Pro and Gemini-2.5-Flash on the MMAU benchmark, as well as Gemini-2.5-Flash and GPT-4o-Audio on MMSU. These results demonstrate the powerful capabilities of \method in general audio understanding and reasoning.
\input{posttrain_tables/voicechat.tex}
% \input{posttrain_tables/audio_reasoning.tex}

% with outstanding scores of 52.0 and 52.1 
For music understanding, we compare \method-Instruct with both generalist audio language models and specialist models in Table \ref{tab:audio_music}. For multi-label classification tasks on MTG-Jamendo and MagnaTagATune, we use micro F1 to compare with BERT-like music specialists instead of AP/AUROC, as language models output discrete label sets without calibrated per-label probabilities/scores required by ranking-based metrics. It is shown in Table \ref{tab:audio_music} that \method-Instruct achieve state-of-the-art performance on RUL-MuchoMusic. On GTZAN, MTG-Jamendo, and MagnaTagATune, the scores of \method-Instruct also significantly surpass other audio language models, including Gemini-2.5-Pro and GPT-4o-Audio, as well as self-supervised music specialist models probed on the respective datasets. These results demonstrate the superior capabilities of \method-Instruct across a variety of music understanding tasks.
\input{posttrain_tables/audio_music.tex}

% \vspace{-.1in}

% Baichuan-Omni-1.5\citep{li2025baichuan}
% MiniCPM-o~\citep{yao2024minicpm}
% Lyra-Base~\citep{zhong2024lyra}
% Megrez-3B-Omni ~\citep{Megrez-3B-Omni}
% VITA-1.5 ~\citep{fu2025vita}
\subsubsection{Performance of Vision $\to$ Text}
To comprehensively evaluate the capabilities on Vision $\to$ Text, we compare \method-Instruct with the Qwen2.5-VL-72B and other good-performing closed-source vision-language models. As illustrated in Table \ref{tab:qwen3-omni-vl-results}, \method-Instruct demonstrates comparable performance to Qwen2.5-VL-72B, and attains better results on Math \& STEM related tasks like MMMU-Pro\textsubscript{overall}, MathVista\textsubscript{mini}, and MATH-Vision\textsubscript{full},  than other vision language models including GPT4-o and Gemini-2.0-Flash. These results reveal the excellent capability of our model on image understanding and reasoning tasks. 

To assess its capabilities, we evaluated the performance of \method-Thinking against several state-of-the-art reasoning models. The comparative results, summarized in Table \ref{tab:qwen3-omni-vl-reasoning-results}, indicate that our proposed model achieves significant advancements. For instance, on Math and STEM benchmarks, it outperforms the \method-Instruct baseline by 4.4 points. It is also noteworthy that our \method-30B-A3B-Thinking model attains a performance level on par with substantially larger baselines, which highlights its excellent balance of effectiveness and computational efficiency.
% Of particular note, \method-30B-A3B-Thinking achieves performance comparable to the much larger InternVL-3.5-241B-A28B model, despite possessing only one-eighth of its parameters, highlighting its remarkable efficiency. 
A limitation of the current model is its suboptimal performance on long video benchmarks. This deficiency stems from two architectural constraints: a limited capacity for positional extrapolation and a restricted context length. Addressing these constraints is a key objective for future work.

\vspace{-.1in}

\input{posttrain_tables/vl}

% For visual grounding, we compare \method with Qwen2.5-VL-7B and other leading LVLMs including Gemini and Grounding-DINO~\citep{liu2024groundingdinomarryingdino}. As illustrated in Table \ref{tab:qwen3-omni-vl-results}, our model outperforms other models across most benchmarks from box-grounding to point-grounding and achieves a good performance of 42.2mAP on open-vocabulary object detection, which reveals the strong visual grounding capability of our model.

% Similar to Image$\to$Text, we compare \method with Qwen2.5-VL-7B and other omni models. As shown in Table \ref{tab:qwen3-omni-vl-results}, \method outperforms all other state-of-the-art open-sourced omni models, and attains better or competitive results compared to Qwen2.5-VL-72B, which demonstrates the superior performance on video understanding.
% \input{posttrain_tables/vl}

\subsubsection{Performance of AudioVisual Video$\to$Text}
As is shown in Table \ref{tab:qwen3-omni-va-results}, the experimental results validate the efficacy of \method across diverse audiovisual tasks. For general understanding, \method-Instruct achieves state-of-the-art performance on the WorldSense benchmark, surpassing other Omni models by a substantial margin. This outcome demonstrates its effectiveness in foundational multimodal integration. Moreover, the model exhibits enhanced performance on complex reasoning tasks, as illustrated in Table \ref{tab:qwen3-omni-va-reasoning-results}, particularly on benchmarks that necessitate reasoning over interconnected audio and visual information. These findings collectively suggest that \method possesses considerable potential for advanced perception and reasoning in real-world contexts.
% DailyOmni, WorldSense, and VideoHolmes benchmarks, surpassing other Omni models by a large margin, which demonstrates the superiority of our model in multimodality understanding.
\input{posttrain_tables/va}

\subsection{Evaluation of X$\to$Speech}
In this section, we evaluate the speech generation capabilities of \method. Due to the lack of relevant assessments, the evaluation of speech generation focuses primarily speech generation given texts, similarity to text-to-speech (TTS), on following three aspects:

\begin{itemize}
    \item \textbf{Zero-Shot Speech Generation}: We assess the content consistency (WER) and speaker similarity (SIM) of our model in zero-shot speech generation on SEED~\citep{seedtts}.
    \item \textbf{Multilingual Speech Generation}: We assess the content consistency and speaker similarity of our model in zero-shot multilingual speech generation on MiniMax multilingual test set~\citep{minimaxspeech}.
    \item \textbf{Cross-Lingual Speech Generation}: We assess the content consistency of our model in zero-shot cross-lingual speech generation on CV3-Eval~\citep{cosyvoice3}.
    % \item \textbf{Single-Speaker Speech Generation}. We assess the stability of our speaker fine-tuned model on the SEED~\citep{seedtts} and MiniMax multilingual test set~\citep{minimaxspeech}.
\end{itemize}

\subsubsection{Evaluation of Zero-Shot Speech Generation}
We compare the \method with state-of-the-art zero-shot TTS systems. As shown in Table~\ref{tab:zero_shot_speech_generation_table}, \method demonstrates highly competitive performance, highlighting its robust speech understanding and generation capabilities developed through pretraining and continual pretraining. Additionally, with reinforcement learning (RL) optimization, \method yields significant improvements in generation stability, which achieves the best performance in the test-en set.

\vspace{-.1in}

\begin{table}[H]
\centering
\caption{\textbf{Zero-Shot Speech Generation on Seed-TTS Test Set. The highest scores are shown in bold.}}
\vspace{-.1in}
\setlength{\tabcolsep}{2.6pt}
\begin{tabular}{@{}cll@{}}
\toprule
\textbf{Datasets} & \textbf{Model} & \textbf{Performance} \\
\midrule
\multicolumn{3}{c}{\textit{Content Consistency}} \\
\midrule 
\multirow{9}{*}{\begin{tabular}[c]{@{}c@{}}\textbf{SEED} \\ \textit{test-zh} | \textit{test-en} \end{tabular}}   
   & Seed-TTS\textsubscript{ICL}~\citep{seedtts} & 1.11 | 2.24  \\ 
   & Seed-TTS\textsubscript{RL}~\citep{seedtts}  & 1.00 | 1.94  \\ 
   & MaskGCT~\citep{maskgct}                     & 2.27 | 2.62  \\ 
   & E2 TTS~\citep{e2tts}                        & 1.97 | 2.19  \\ 
   & F5-TTS~\citep{f5tts}                        & 1.56 | 1.83  \\ 
   & Spark TTS~\citep{sparktts}            & 1.20 | 1.98  \\
   & CosyVoice 2~\citep{cosyvoice2}              & 1.45 | 2.57  \\
   & CosyVoice 3~\citep{cosyvoice3}            & \textbf{0.71} | 1.45  \\
   % & Qwen2.5-Omni-7B\textsubscript{ICL}~\citep{qwen2.5omni}                & 1.70 | 2.72 | 7.97 \\
   & Qwen2.5-Omni-7B~\citep{qwen2.5omni}                 & 1.42 | 2.33 \\
   % & \method-30B-A3B\textsubscript{ICL}                & - | - | - \\
   & \method-30B-A3B                 & 1.07 | \textbf{1.39} \\
\bottomrule
\end{tabular}
\label{tab:zero_shot_speech_generation_table}
\end{table}

\subsubsection{Evaluation of Multilingual Speech Generation}
\method supports speech generation across 10 languages. We evaluate its performance against both the MiniMax-Speech and ElevenLabs Multilingual v2 models for multilingual speech generation. As shown in Table~\ref{tab:multilingual_speech_generation_table}, \method surpasses these models by a significant margin for languages such as Chinese, English, and French, while delivering competitive results in the remaining languages. These findings indicate that \method generates cloned speech with consistent  stability and human-like voice across all evaluated languages.

% the performance of  \method  is comparable to that of  MiniMax-Speech and ElevenLabs Multilingual v2 model

% \usepackage{booktabs}
% \usepackage{multirow}
\begin{table}[H]
\centering
\caption{\textbf{Multilingual Speech Generation on MiniMax Multilingual Test Set. The highest scores are shown in bold.}}
\begin{tabular}{@{}lllllll@{}}
\toprule
\multirow{2}{*}{\textbf{Language}} & \multicolumn{3}{c}{\textbf{Content Consistency}}      & \multicolumn{3}{c}{\textbf{Speaker Similarity}}           \\ \cmidrule(l){2-7} 
                          & \tabincell{c}{\textbf{\method}\\\textbf{-30B-A3B}}       & \textbf{MiniMax} & \textbf{ElevenLabs} & \tabincell{c}{\textbf{\method}\\\textbf{-30B-A3B}}       & \textbf{MiniMax} & \textbf{ElevenLabs} \\ \midrule
Chinese                    & \textbf{0.716} & 2.252   & 16.026      & 0.772          & \textbf{0.780}   & 0.677      \\
English                    & \textbf{1.069} & 2.164   & 2.339      & \textbf{0.773}          & 0.756   & 0.613      \\
German                    & 0.777 & 1.906   & \textbf{0.572}      & \textbf{0.738}          & 0.733   & 0.614      \\
Italian                   & \textbf{1.067} & 1.543   & 1.743      & \textbf{0.742} & 0.699   & 0.579      \\
Portuguese                & 1.872          & 1.877   & \textbf{1.331}      & 0.770          & \textbf{0.805}   & 0.711      \\
Spanish                   & 1.765          & \textbf{1.029}   & 1.084      & 0.744 & \textbf{0.762}   & 0.615      \\
Japanese                  & 3.631          & \textbf{3.519}   & 10.646     & 0.763 & \textbf{0.776}   & 0.738      \\
Korean                    & \textbf{1.670}          & 1.747   & 1.865      & \textbf{0.778} & 0.776   & 0.700      \\
French                    & \textbf{2.505} & 4.099   & 5.216      & \textbf{0.689} & 0.628   & 0.535      \\
Russian                   & 3.986          & 4.281   & \textbf{3.878}      & 0.759          & \textbf{0.761}   & 0.676      \\ \bottomrule
% Thai                      & 5.437          & \textbf{2.701}   & 73.936     & 0.771          & \textbf{0.800}   & 0.588      \\
% Indonesian                & 1.219          & 1.237   & \textbf{1.059}      & \textbf{0.733} & 0.729   & 0.660      \\ \bottomrule
\end{tabular}
\label{tab:multilingual_speech_generation_table}
\end{table}

% depends on the performance
\subsubsection{Evaluation of Cross-Lingual Speech Generation}
\method supports not only multilingual voice cloning but also cross-lingual voice cloning. We evaluate its performance against CosyVoice2 and CosyVoice3 for cross-lingual speech generation. As shown in Table~\ref{tab:cross_lingual_speech_generation_table}, \method outperforms CosyVoice3 in any-to-en (any language to English) and any-to-ko (any language to Korean) voice cloning. Notably, in any-to-ja (any language to Japanese) tasks, \method achieves comparable performance to CosyVoice3 even without text normalization, despite CosyVoice3 converting all Japanese characters into phonetic kana. These results highlight \method's superiority in cross-lingual speech generation, demonstrating its adaptability across diverse linguistic contexts.

% \usepackage{booktabs}
\begin{table}[H]
\centering
\caption{\textbf{Cross-Lingual Speech Generation on CosyVoice3 Cross-Lingual Test Set. The highest scores are shown in bold.}}
\begin{tabular}{@{}lccc@{}}
\toprule
\textbf{Language} & \multicolumn{1}{c}{\textbf{\method-30B-A3B}} & \multicolumn{1}{c}{\textbf{CosyVoice3}} & \multicolumn{1}{c}{\textbf{CosyVoice2}} \\ \midrule
en-to-zh & 5.37                        & \textbf{5.09}                           & 13.5                           \\
ja-to-zh & 3.32                        & \textbf{3.05}                           & 48.1                           \\
ko-to-zh & \textbf{0.99}                        & 1.06                           & 7.70                           \\
zh-to-en & \textbf{2.76}                        & 2.98                           & 6.47                           \\
ja-to-en & \textbf{3.31}                        & 4.20                           & 17.1                           \\
ko-to-en & \textbf{3.34}                        & 4.19                           & 11.2                           \\
zh-to-ja & 8.29                        & \textbf{7.08}                           & 13.1                           \\
en-to-ja & 7.53                        & \textbf{6.80}                           & 14.9                           \\
ko-to-ja & 4.24                        & \textbf{3.93}                           & 5.86                           \\
zh-to-ko & \textbf{5.13}                        & 14.4                           & 24.8                           \\
en-to-ko & \textbf{4.96}                        & 5.87                           & 21.9                           \\
ja-to-ko & \textbf{6.23}                        & 7.92                           & 21.5                           \\ \bottomrule
\end{tabular}
\label{tab:cross_lingual_speech_generation_table}
\end{table}

%% file: posttrain_tables/text.tex
% \usepackage{booktabs}
% \usepackage{multirow}
% \usepackage{graphicx}
\begin{table}[H]
\centering
\caption{\textbf{Text $\to$ Text performance of \method-Instruct and other non-reasoning baselines. The highest
scores are shown in bold.}}
\vspace{-.1in}
\label{tab:text_nonthink}
\resizebox{\textwidth}{!}{%
\begin{tabular}{@{}clccccc@{}}
\toprule
\multicolumn{1}{l}{\textit{}} &
  \textit{} &
  \textbf{GPT-4o-0327} &
  \begin{tabular}[c]{@{}c@{}}\textbf{Qwen3-235B-A22B}\\ \textbf{Non Thinking}\end{tabular} &
  \begin{tabular}[c]{@{}c@{}}\textbf{Qwen3-30B-A3B}\\ \textbf{-Instruct-2507}\end{tabular} &
  \begin{tabular}[c]{@{}c@{}}\textbf{Qwen3-Omni-30B-A3B}\\ \textbf{-Instruct}\end{tabular} &
  \begin{tabular}[c]{@{}c@{}}\textbf{Qwen3-Omni-Flash}\\ \textbf{-Instruct}\end{tabular} \\ \midrule
\multirow{2}{*}{GeneralTasks}                                                 & MMLU-Redux          & \textbf{91.3} & 89.2          & 89.3          & 86.6 & 86.8          \\
                                                                              & GPQA                & 66.9          & 62.9          & \textbf{70.4} & 69.6 & 69.7          \\ \midrule
\multirow{2}{*}{Reasoning}                                                    & AIME25              & 26.7          & 24.7          & 61.3          & 65.0 & \textbf{65.9} \\
                                                                              & ZebraLogic          & 52.6          & 37.7          & \textbf{90.0} & 76.0 & 76.1          \\ \midrule
Code                                                                          & MultiPL-E           & 82.7          & 79.3          & \textbf{83.8} & 81.4 & 81.5          \\ \midrule
\multirow{3}{*}{\begin{tabular}[c]{@{}c@{}}Alignment\\ Tasks\end{tabular}}    & IFEval              & 83.9          & 83.2          & \textbf{84.7} & 81.0 & 81.7          \\
                                                                              & Creative Writing v3 & 84.9          & 80.4          & \textbf{86.0} & 80.6 & 81.8          \\
                                                                              & WritingBench        & 75.5          & 77.0          & \textbf{85.5} & 82.6 & 83.0          \\ \midrule
Agent                                                                         & BFCL-v3             & 66.5          & \textbf{68.0} & 65.1          & 64.4 & 65.0          \\ \midrule
\multirow{2}{*}{\begin{tabular}[c]{@{}c@{}}Multilingual\\ Tasks\end{tabular}} & MultiIF             & \textbf{70.4} & 70.2          & 67.9          & 64.0 & 64.7          \\
                                                                              & PolyMATH            & 25.5          & 27.0          & \textbf{43.1} & 37.9 & 39.3          \\ \bottomrule
\end{tabular}%
}
\end{table}

\vspace{-0.1cm}

% \usepackage{booktabs}
% \usepackage{multirow}
% \usepackage{graphicx}
\begin{table}[H]
\centering
\caption{\textbf{Text $\to$ Text performance of Qwen3-Omni-Thinking and other reasoning baselines. The highest
scores are shown in bold.}}
\vspace{-0.1in}
\label{tab:text_think}
\resizebox{\textwidth}{!}{%
\begin{tabular}{@{}clccccc@{}}
\toprule
\multicolumn{1}{l}{\textit{}} & \textit{} & 
\begin{tabular}[c]{@{}c@{}}\textbf{Gemini-2.5-Flash} \\ \textbf{Thinking}\end{tabular} & 
\begin{tabular}[c]{@{}c@{}}\textbf{Qwen3-235B-A22B}\\ \textbf{Thinking}\end{tabular} & 
\begin{tabular}[c]{@{}c@{}}\textbf{Qwen3-30B-A3B}\\ \textbf{-Thinking-2507}\end{tabular} & 
\begin{tabular}[c]{@{}c@{}}\textbf{Qwen3-Omni-30B-A3B}\\ \textbf{-Thinking}\end{tabular} & 
\begin{tabular}[c]{@{}c@{}}\textbf{Qwen3-Omni-Flash}\\ \textbf{-Thinking}\end{tabular} \\ \midrule
\multirow{2}{*}{\textit{\begin{tabular}[c]{@{}c@{}}General\\ Tasks\end{tabular}}} & MMLU-Redux & 92.1 & \textbf{92.7} & 91.4 & 88.8 & 89.7 \\
 & GPQA & \textbf{82.8} & 71.1 & 73.4 & 73.1 & 73.1 \\ \midrule
\multirow{2}{*}{\textit{Reasoning}} & AIME25 & 72.0 & 81.5 & \textbf{85.0} & 73.7 & 74.0 \\
 & LiveBench 20241125 & 74.3 & \textbf{77.1} & 76.8 & 71.8 & 70.3 \\ \midrule
\textit{Code} & MultiPL-E & \textbf{84.5} & 79.9 & 81.3 & 80.6 & 81.0 \\ \midrule
\multirow{4}{*}{\textit{\begin{tabular}[c]{@{}c@{}}Alignment\\ Tasks\end{tabular}}} & IFEval & \textbf{89.8} & 83.4 & 88.9 & 85.1 & 85.2 \\
 & Arena-Hard v2 & 56.7 & \textbf{61.5} & 56.0 & 55.1 & 57.8 \\
 & Creative Writing v3 & \textbf{85.0} & 84.6 & 84.4 & 82.5 & 83.6 \\
 & WritingBench & 83.9 & 80.3 & 85.0 & 85.5 & \textbf{85.9} \\ \midrule
\textit{Agent} & BFCL-v3 & 68.6 & 70.8 & \textbf{72.4} & 63.2 & 64.5 \\ \midrule
\multirow{2}{*}{\textit{\begin{tabular}[c]{@{}c@{}}Multilingual\\ Tasks\end{tabular}}} & MultiIF & 74.4 & 71.9 & \textbf{76.4} & 72.9 & 73.2 \\
 & PolyMATH & 49.8 & \textbf{54.7} & 52.6 & 47.1 & 48.7 \\ \bottomrule
\end{tabular}%
}
\end{table}

%% file: posttrain_tables/audio_asrs2tt.tex
% \usepackage{booktabs}
% \usepackage{graphicx}
\begin{table}[H]
\centering
\caption{\textbf{Transcription performance for Audio$\to$Text tasks (ASR \& S2TT), comparing Qwen3-Omni-Instruct with the baselines. The highest scores are shown in bold.}}
\vspace{-0.1in}
\label{tab:audio_asr}
\resizebox{\textwidth}{!}{%
\begin{threeparttable}
\begin{tabular}{@{}lcccccccc@{}}
\toprule
 &
  \begin{tabular}[c]{@{}c@{}}\textbf{Seed}\\ \textbf{-ASR}\end{tabular} &
\begin{tabular}[c]{@{}c@{}}\textbf{Voxtral}\\ \textbf{-Mini}\end{tabular} &
\begin{tabular}[c]{@{}c@{}}\textbf{Voxtral}\\ \textbf{-Small}\end{tabular} &
\begin{tabular}[c]{@{}c@{}}\textbf{GPT-4o}\\ \textbf{-Transcribe}\end{tabular} &
\begin{tabular}[c]{@{}c@{}}\textbf{Gemini-2.5}\\ \textbf{-Pro}\end{tabular} &
\begin{tabular}[c]{@{}c@{}}\textbf{Qwen2.5}\\ \textbf{-Omni}\end{tabular} &
\begin{tabular}[c]{@{}c@{}}\textbf{Qwen3-Omni}\\ \textbf{-30B-A3B-Instruct}\end{tabular} &
\begin{tabular}[c]{@{}c@{}}\textbf{Qwen3-Omni}\\ \textbf{-Flash-Instruct}\end{tabular} \\ \midrule
\multicolumn{9}{c}{\textit{EN \& ZH ASR (wer)}} \\ \midrule
\begin{tabular}[c]{@{}l@{}}Wenetspeech\\ \textit{net}|\textit{meeting}\end{tabular} &
  4.66|\textbf{5.69} &
  24.30|31.53 &
  20.33|26.08 &
  15.30|32.27 &
  14.43|13.47 &
  5.91|7.65 &
  4.69|5.89 &
  \textbf{4.62}|5.75 \\
\begin{tabular}[c]{@{}l@{}}Librispeech\\ \textit{clean} | \textit{other}\end{tabular} &
  1.58|2.84 &
  1.88|4.12 &
  1.56|3.30 &
  1.39|3.75 &
  2.89|3.56 &
  1.74|3.45 &
  \textbf{1.22}|2.48 &
  1.27|\textbf{2.44} \\
CV15-en &
  - &
  9.47 &
  7.79 &
  10.01 &
  9.89 &
  7.61 &
  6.05 &
  \textbf{5.94} \\
CV15-zh &
  - &
  24.67 &
  19.30 &
  9.84 &
  8.00 &
  5.13 &
  4.31 &
  \textbf{4.28} \\
Fleurs-en &
  3.40 &
  3.96 &
  3.77 &
  3.32 &
  2.94 &
  3.77 &
  \textbf{2.72} &
  2.74 \\
Fleurs-zh &
  2.69 &
  12.22 &
  7.98 &
  2.44 &
  2.71 &
  2.54 &
  2.20 &
  \textbf{2.19} \\ \midrule
\multicolumn{9}{c}{\textit{Multilingual ASR (wer)}} \\ \midrule
\begin{tabular}[c]{@{}l@{}}Fleurs-avg\\ (19 lang)\tnote{a}\end{tabular} &
  - &
  15.67 &
  8.09 &
  4.48 &
  5.55 &
  14.04 &
  5.33 &
  \textbf{5.31} \\ \midrule
\multicolumn{9}{c}{\textit{Lyric ASR (wer)}} \\ \midrule
MIR-1K (vocal-only)\tnote{b} &
  6.45 &
  23.33 &
  18.73 &
  11.87 &
  9.85 &
  8.15 &
  5.90 &
  \textbf{5.85} \\
Opencpop-test &
  2.98 &
  31.01 &
  16.06 &
  7.93 &
  6.49 &
  2.84 &
  \textbf{1.54} &
  2.02 \\ \midrule
\multicolumn{9}{c}{\textit{S2TT (BLEU)}} \\ \midrule
Fleurs-en2xx\tnote{c} &
  - &
  30.35 &
  37.85 &
  - &
  \textbf{39.25} &
  29.22 &
  37.50 &
  36.22 \\
Fleurs-xx2en &
  - &
  27.54 &
  32.81 &
  - &
  \textbf{35.41} &
  28.61 &
  31.08 &
  30.71 \\
Fleurs-zh2xx &
  - &
  17.03 &
  22.05 &
  - &
  \textbf{26.63} &
  17.97 &
  25.17 &
  25.10 \\
Fleurs-xx2zh &
  - &
  28.75 &
  34.82 &
  - &
  \textbf{37.50} &
  27.68 &
  33.13 &
  31.19 \\ \bottomrule
\end{tabular}%
\begin{tablenotes} 
    \item[a] These 19 languages include Arabic, Cantonese, Chinese, Dutch, English, French, German, Indonesian, Italian, Japanese, Korean, Malay, Portuguese, Russian, Spanish, Thai, Turkish, Urdu, Vietnamese.
    \item[b] Transcription is converted into Simplified Chinese.
    \item[c] The results encompass translations across 15 languages: Arabic, Cantonese, Chinese, English, French, German, Indonesian, Italian, Japanese, Korean, Portuguese, Russian, Spanish, Thai, Vietnamese. For notation, ``en2xx'' denotes translation from English into each of the other 14 target languages, where ``xx'' ranges over the remaining language codes.
\end{tablenotes}
\end{threeparttable}
}
\end{table}

%% file: posttrain_tables/voicechat.tex
% \usepackage{booktabs}
% \usepackage{graphicx}
\begin{table}[H]
\centering
\caption{\textbf{Voice interaction and audio reasoning performance for Audio$\to$Text tasks, comparing Qwen3-Omni with the baselines. The highest scores are shown in bold.}}
\label{tab:audio_voicebench}
\resizebox{\textwidth}{!}{%
\begin{tabular}{@{}lcccccccc@{}}
\toprule
 &
  \begin{tabular}[c]{@{}c@{}}\textbf{GPT-4o} \\ \textbf{-Audio}\end{tabular} &
\begin{tabular}[c]{@{}c@{}}\textbf{Gemini-2.5} \\ \textbf{-Flash}\end{tabular} &
\begin{tabular}[c]{@{}c@{}}\textbf{Gemini-2.5} \\ \textbf{-Pro}\end{tabular} &
\begin{tabular}[c]{@{}c@{}}\textbf{Qwen2.5} \\ \textbf{-Omni}\end{tabular} &
\begin{tabular}[c]{@{}c@{}}\textbf{Qwen3-Omni} \\ \textbf{ -30B-A3B-Instruct}\end{tabular} &
\begin{tabular}[c]{@{}c@{}}\textbf{Qwen3-Omni} \\ \textbf{-30B-A3B-Thinking}\end{tabular} &
\begin{tabular}[c]{@{}c@{}}\textbf{Qwen3-Omni} \\ \textbf{-Flash-Instruct}\end{tabular} &
\begin{tabular}[c]{@{}c@{}}\textbf{Qwen3-Omni} \\ \textbf{-Flash-Thinking}\end{tabular} \\ \midrule
\multicolumn{9}{c}{\textit{VoiceBench}}                                                                             \\ \midrule
AlpacaEval     & 95.6 & 96.1 & 94.3          & 89.9 & 94.8          & 96.4 & 95.4          & \textbf{96.8} \\
CommonEval     & 89.8 & 88.3 & 88.4          & 76.7 & 90.8          & 90.5 & \textbf{91.0} & 90.9          \\
WildVoice      & 91.6 & 92.1 & 93.4          & 77.7 & 91.6          & 90.5 & \textbf{92.3} & 90.9          \\
SD-QA          & 75.5 & 84.5 & \textbf{90.1} & 56.4 & 76.9          & 78.1 & 76.8          & 78.5          \\
MMSU           & 80.3 & 66.1 & 71.1          & 61.7 & 68.1          & 83.0 & 68.4          & \textbf{84.3} \\
OpenBookQA     & 89.2 & 56.9 & 92.3          & 80.9 & 89.7          & 94.3 & 91.4          & \textbf{95.0} \\
BBH            & 84.1 & 83.9 & \textbf{92.6} & 66.7 & 80.4          & 88.9 & 80.6          & 89.6          \\
IFEval         & 76.0 & 83.8 & \textbf{85.7} & 53.5 & 77.8          & 80.6 & 75.2          & 80.8          \\
AdvBench       & 98.7 & 98.9 & 98.1          & 99.2 & \textbf{99.3} & 97.2 & \textbf{99.4} & 98.9          \\
Overall        & 86.8 & 83.4 & \textbf{89.6} & 73.6 & 85.5          & 88.8 & 85.6          & 89.5          \\ \midrule
\multicolumn{9}{c}{\textit{Audio Reasoning}}                                                                        \\ \midrule
MMAU-v05.15.25 & 62.5 & 71.8 & 77.4          & 65.5 & 77.5          & 75.4 & \textbf{77.6} & 76.5          \\
MMSU           & 56.4 & 70.2 & \textbf{77.7} & 62.6 & 69.0          & 70.2 & 69.1          & 71.3          \\ \bottomrule
\end{tabular}%
}
\end{table}

%% file: posttrain_tables/audio_music.tex
% \usepackage{booktabs}
% \usepackage{graphicx}
\begin{table}[H]
\centering
\caption{\textbf{Music understanding performance for Audio$\to$Text tasks, comparing Qwen3-Omni-Instruct with baselines. The highest scores are shown in bold.}}
\vspace{-0.1in}
\label{tab:audio_music}
\resizebox{\textwidth}{!}{%
\begin{tabular}{@{}lcccccc@{}}
\toprule
 &
  \begin{tabular}[c]{@{}c@{}}\textbf{Best Specialist} \\ \textbf{Models}\end{tabular} &
  \begin{tabular}[c]{@{}c@{}}\textbf{GPT-4o} \\ \textbf{-Audio}\end{tabular} &
  \begin{tabular}[c]{@{}c@{}}\textbf{Gemini-2.5} \\ \textbf{-Pro}\end{tabular} &
  \begin{tabular}[c]{@{}c@{}}\textbf{Qwen2.5} \\ \textbf{-Omni}\end{tabular} &
  \begin{tabular}[c]{@{}c@{}}\textbf{Qwen3-Omni} \\ \textbf{-30B-A3B-Instruct}\end{tabular} &
  \begin{tabular}[c]{@{}c@{}}\textbf{Qwen3-Omni} \\ \textbf{-Flash-Instruct}\end{tabular} \\ \midrule
RUL-MuchoMusic &
  \begin{tabular}[c]{@{}c@{}}47.6 (Audio Flamingo 3)\\ \citep{goel2025audio}\end{tabular} &
  36.1 &
  49.4 &
  47.3 &
  52.0 &
  \textbf{52.1} \\
\begin{tabular}[c]{@{}l@{}}GTZAN\\ \textit{Acc.} \end{tabular} &
  \begin{tabular}[c]{@{}c@{}}87.9 (CLaMP 3)\\ \citep{wu2025clamp}\end{tabular} &
  76.5 &
  81.0 &
  81.7 &
  93.0 &
  \textbf{93.1} \\
\begin{tabular}[c]{@{}l@{}}MTG Genre\\ \textit{Micro F1}\end{tabular} &
  \begin{tabular}[c]{@{}c@{}}35.8 (MuQ-MuLan)\\ \citep{zhu2025muq}\end{tabular} &
  25.3 &
  32.6 &
  32.5 &
  39.0 &
  \textbf{39.5} \\
\begin{tabular}[c]{@{}l@{}}MTG Mood/Theme\\ \textit{Micro F1}\end{tabular} &
  \begin{tabular}[c]{@{}c@{}}10.9 (MuQ-MuLan)\\ \citep{zhu2025muq}\end{tabular} &
  11.3 &
  14.1 &
  8.9 &
  21.0 &
  \textbf{21.7} \\
\begin{tabular}[c]{@{}l@{}}MTG Instrument\\ \textit{Micro F1}\end{tabular} &
  \begin{tabular}[c]{@{}c@{}}39.8 (MuQ-MuLan)\\ \citep{zhu2025muq}\end{tabular} &
  34.2 &
  33.0 &
  22.6 &
  40.5 &
  \textbf{40.7} \\
\begin{tabular}[c]{@{}l@{}}MTG Top50\\ \textit{Micro F1}\end{tabular} &
  \begin{tabular}[c]{@{}c@{}}33.2 (MuQ-MuLan)\\ \citep{zhu2025muq}\end{tabular} &
  25.0 &
  26.1 &
  21.6 &
  36.7 &
  \textbf{36.9} \\
\begin{tabular}[c]{@{}l@{}}MagnaTagATune\\ \textit{Micro F1}\end{tabular} &
  \begin{tabular}[c]{@{}c@{}}41.6 (MuQ)\\ \citep{zhu2025muq}\end{tabular} &
  29.2 &
  28.1 &
  30.1 &
  44.3 &
  \textbf{46.8} \\ \bottomrule
\end{tabular}%
}
\end{table}

%% file: posttrain_tables/vl.tex
\begin{table}[H]
\centering
\caption{\textbf{Vision $\to$ Text performance of \method-Instruct and other non-reasoning baselines. The highest scores are shown in bold.}}
\vspace{-.1in}
\label{tab:qwen3-omni-vl-results}
% \adjustbox{center=\textwidth}{
% \small
% \setlength{\tabcolsep}{3pt} %
\resizebox{\textwidth}{!}{%
\begin{tabular}{@{}lccccc@{}}
\toprule
\textbf{Datasets} & \tabincell{c}{\textbf{GPT4-o}} & \tabincell{c}{\textbf{Gemini-2.0-Flash}} & \tabincell{c}{\textbf{Qwen2.5-VL}\\\textbf{72B}} & \tabincell{c}{\textbf{\method-30B-A3B}\\\textbf{-Instruct}} & \tabincell{c}{\textbf{\method-Flash}\\\textbf{-Instruct}} \\
\midrule
\multicolumn{6}{c}{\textit{General Visual Question Answering}} \\
\midrule
MMStar & 64.7 & \textbf{71.4} & 70.8 & 68.5 & 69.3 \\
HallusionBench & 55.0 & 56.3 & 55.2 & 59.7 & \textbf{60.4} \\
MM-MT-Bench & \textbf{7.7} & 6.7 & 7.6 & 7.4 & 7.6 \\
\midrule
\multicolumn{6}{c}{\textit{Math \& STEM}} \\
\midrule
MMMU\textsubscript{val}  & 69.1 & \textbf{71.3} & 70.2 & 69.1 & 69.8 \\
MMMU-Pro\textsubscript{overall} & 51.9 & 56.1 & 51.1 & 57.0 & \textbf{58.2} \\
MathVista\textsubscript{mini} & 63.8 & 71.4 & 74.8 & 75.9 & \textbf{77.4} \\
MATH-Vision\textsubscript{full} & 30.4 & 48.6 & 38.1 & 56.3 & \textbf{57.3} \\
\midrule
\multicolumn{6}{c}{\textit{Documentation Understanding}} \\
\midrule
AI2D\textsubscript{w.M.} & 84.6 & 86.7 & \textbf{88.7} & 85.2 & 86.4 \\
ChartQA\textsubscript{test Avg.} & 86.7 & 64.6 & \textbf{89.5} & 86.8  & 87.1 \\
\midrule
\multicolumn{6}{c}{Counting} \\
\midrule
CountBench & 87.9 & 91.2 & \textbf{93.6} & 90.0 & 90.0 \\
\midrule
\multicolumn{6}{c}{Video Understanding} \\
\midrule
Video-MME\textsubscript{w/o sub} & 71.9 & 72.4 & \textbf{73.3} & 70.5 & 71.4 \\
LVBench & 30.8 & \textbf{57.9} & 47.3 & 50.2 & 51.1 \\
MLVU & 64.6 & 71.0 & 74.6 & 75.2 & \textbf{75.7} \\
\bottomrule
\end{tabular}
}
\end{table}

\vspace{-.15in}

\begin{table}[H]
\centering
\caption{\textbf{Vision $\to$ Text performance of \method-Thinking and other reasoning baselines. The highest
scores are shown in bold.}}

\vspace{-.1in}
\label{tab:qwen3-omni-vl-reasoning-results}
\resizebox{\textwidth}{!}{%
% \adjustbox{center=\textwidth}{
% \small
% \setlength{\tabcolsep}{3pt} %
\begin{tabular}{@{}lcccc@{}}
\toprule
\textbf{Datasets}  & \tabincell{c}{\textbf{Gemini-2.5-Flash}\\\textbf{-Thinking}} & \tabincell{c}{\textbf{InternVL-3.5-241B-A28B}} & \tabincell{c}{\textbf{\method-30B-A3B}\\\textbf{-Thinking}} & \tabincell{c}{\textbf{\method-Flash}\\\textbf{-Thinking}} \\
\midrule
\multicolumn{5}{c}{\textit{General Visual Question Answering}} \\
\midrule
MMStar & 75.5 & \textbf{77.9} & 74.9 & 75.5 \\
HallusionBench & 61.1 & 57.3 & 62.8 & \textbf{63.4} \\
MM-MT-Bench & 7.8 & -- & \textbf{8.0} & \textbf{8.0} \\
\midrule
\multicolumn{5}{c}{\textit{Math \& STEM}} \\
\midrule
MMMU\textsubscript{val} & 76.9 & \textbf{77.7} & 75.6 & 75.0 \\
MMMU-Pro\textsubscript{overall} & \textbf{65.8} & -- & 60.5 & 60.8 \\
MathVista\textsubscript{mini} & 77.6 & \textbf{82.7} & 80.0 & 81.2 \\
MATH-Vision\textsubscript{full} & 62.3 & \textbf{63.9} & 62.9 & 63.8 \\
\midrule
\multicolumn{5}{c}{\textit{Documentation Understanding}} \\
\midrule
AI2D\textsubscript{w.M.} & \textbf{88.6} & 87.3 & 86.1 & 86.8 \\
ChartQA\textsubscript{test Avg.} & -- & 88.0 & \textbf{89.5} & 89.3 \\
\midrule
\multicolumn{5}{c}{\textit{Counting}} \\
\midrule
CountBench & 88.6 & -- & 88.6 & \textbf{92.5} \\
\midrule
\multicolumn{5}{c}{\textit{Video Understanding}} \\
\midrule
Video-MME\textsubscript{w/o sub} & \textbf{79.6} & 72.9 & 69.7 & 69.8 \\
LVBench & \textbf{64.5} & -- & 49.0 & 49.5 \\
MLVU & \textbf{82.1} & 78.2 & 72.9 & 73.9 \\
\bottomrule
\end{tabular}
}
\end{table}

%% file: posttrain_tables/va.tex
\begin{table}[H]
\centering
\caption{\textbf{AudioVisual $\to$ Text performance of \method-Instruct and other non-reasoning baselines. The highest
scores are shown in bold.}}
\vspace{-1mm}
\label{tab:qwen3-omni-va-results}
% \adjustbox{center=\textwidth}{
% \small
% \setlength{\tabcolsep}{3pt} %
\resizebox{\textwidth}{!}{%
\begin{tabular}{@{}clcccccc@{}}
\toprule
% & \textbf{Datasets} & \tabincell{c}{\textbf{Gemini-2.5-Flash}}  & \tabincell{c}{\textbf{Prior SoTA}} &\tabincell{c}{\textbf{Qwen2.5-Omni}}  & \tabincell{c}{\textbf{\method-30B-A3B}\\\textbf{-Instruct}}   & \tabincell{c}{\textbf{\method-Flash}\\\textbf{-Instruct}}\\
% \midrule
% & WorldSense & 50.9* &  47.1\citep{humanomniv2} & 45.4 & 54.0 & 54.1  \\
& \textbf{Datasets}  & \tabincell{c}{\textbf{Previous}\\\textbf{Open-source SoTA}} & \tabincell{c}{\textbf{Gemini-2.5-Flash}}  &\tabincell{c}{\textbf{Qwen2.5-Omni}}  & \tabincell{c}{\textbf{\method-30B-A3B}\\\textbf{-Instruct}}   & \tabincell{c}{\textbf{\method-Flash}\\\textbf{-Instruct}}\\
\midrule
& WorldSense &  47.1\citep{humanomniv2}  & 50.9 & 45.4 & 54.0 & \textbf{54.1}  \\
\bottomrule
\end{tabular}
}
\end{table}

\begin{table}[H]
\centering
\caption{\textbf{AudioVisual $\to$ Text performance of Qwen3-Omni-30B-A3B-Thinking and other reasoning baselines. The highest
scores are shown in bold.}}
\vspace{-1mm}
\label{tab:qwen3-omni-va-reasoning-results}
% \adjustbox{center=\textwidth}{
\resizebox{\textwidth}{!}{%
\small
\setlength{\tabcolsep}{3pt} %
\begin{tabular}{@{}clcccccc@{}}
\toprule
& \textbf{Datasets}   & \tabincell{c}{\textbf{Previous}\\\textbf{Open-source SoTA}} & \tabincell{c}{\textbf{Gemini-2.5-Flash}\\\textbf{-Thinking}}  & \tabincell{c}{\textbf{\method-30B-A3B}\\\textbf{-Thinking}}  & \tabincell{c}{\textbf{\method-Flash}\\\textbf{-Thinking}}  \\
\midrule
& DailyOmni   &  69.8\citep{videosalmonn2} & 72.7  & 75.8 & \textbf{76.2}    \\
% & WorldSense  & 52.3* &  47.1 & 52.5 & 53.4  \\
& VideoHolmes  &  55.6\citep{videosalmonn2} & 49.5  & \textbf{57.3} & \textbf{57.3}  \\
\bottomrule
\end{tabular}
}
\end{table}

%% file: content/omniexp.tex
\section{Evaluating Non‑Degradation Across Modalities}\label{sec:omniexp}
% Since the post-training procedures differ across modalities, with varying techniques and optimization goals, a strict and standardized integration of the data is not feasible. To obtain a rigorous conclusion, we pre-trained size-matched text-only, vision-only, and Omni models. The Omni model used exactly the same text and vision datasets as the respective single-modality baselines. We further match the effective number of epochs per modality by tuning sampling ratios and align the learning-rate schedules and batch sizes across models. The only difference is that the Omni model additionally audio data and a small amount of audio-visual video data.
A standardized data integration methodology is rendered impractical by the heterogeneous nature of different modalities, each requiring distinct pre-training objectives and optimization techniques. To ensure a fair and rigorous evaluation, we therefore designed a controlled comparative study. Our approach involved pre-training three models with matched parameter counts: a text-only baseline, a vision-only baseline, and a multimodal ``Omni'' model. To isolate the effects of multimodality, all confounding variables were meticulously controlled. Specifically, the Omni model was trained on the identical text and vision corpora as the unimodal baselines. Moreover, we aligned critical training parameters across all models, including learning rate schedules, batch sizes, and the effective number of training epochs for each modality, which was normalized by adjusting data sampling ratios. Consequently, the sole differentiating factor in our experiment was the Omni model's inclusion of supplementary audio and audio-visual data during its pre-training phase.

The results are shown in Table~\ref{tab:non-degradation across modalities}, we evaluate comprehensive benchmarks covering a variety of modalities, including the text modality (general tasks, math \& STEM tasks, coding tasks, multilingual tasks), the visual modality (college-level problems, OCR-related tasks), and the video modality (video understanding tasks). The experimental results not only demonstrate that mixing unimodal and cross-modal data during the early stage of text pretraining can achieve better performance across all modalities, but also indicate that joint multimodal training enables mutual enhancement between different modalities, leading to improved performance in single modalities as well. 
This fully showcases the versatility and robustness of \method across diverse evaluation criteria.

Due to the prohibitive experimental cost, we could not conduct a comprehensive sweep across all model scales. Based on Table~\ref{tab:non-degradation across modalities} and our internal experiments, we observe: (1) early multimodal integration during pretraining allows language models to be co-trained with vision or audio without any degradation in language capability; (2) the inclusion of the text modality substantially improves performance in the vision and audio. In constrast, we do not observe measurable gains in language ability from adding visual or audio signals; (3) empirically, adding audio data consistently improves vision performance on the MMMU benchmark and OCR-related tasks

\begin{table}[H]
\centering
\caption{\textbf{We compare the performance of 30A3 models that are contemporaneous and identical in size in Qwen series. To ensure experimental rigor, all models were trained under the same schedule, using identical datasets for their respective modalities and exactly matched training compute (FLOPs).}}
\vspace{-1mm}
\label{tab:non-degradation across modalities}
% \adjustbox{center=\textwidth}{
% \small
% \setlength{\tabcolsep}{3pt} %
\resizebox{\textwidth}{!}{%
\begin{tabular}{@{}clccc@{}}
\toprule
& \textbf{Datasets} & \tabincell{c}{\textbf{Qwen3-30B-A3B}\\\textbf{-Base-202507}} & \tabincell{c}{\textbf{Qwen3-VL-30B-A3B}\\\textbf{-Base-202507}} & \tabincell{c}{\textbf{\method-30B-A3B}\\\textbf{-Base-202507}} \\
\midrule
\multirow{4}{*}{\tabincell{c}{\textit{General}\\\textit{Tasks}}} & MMLU & 81.24 & - & \textbf{81.69} \\
& MMLU-Redux & 80.17 & - & \textbf{80.60} \\
& MMLU-Pro & \textbf{61.81} & - & 61.57 \\
& SuperGPQA & 38.24 & - & \textbf{40.14} \\
& BBH & \textbf{83.79} & - & 83.53 \\
\midrule
\multirow{1}{*}{\tabincell{c}{\textit{Math \& STEAM}\\\textit{Tasks}}} & GSM8K & 90.83 & - & \textbf{91.36} \\
& MATH & \textbf{60.84} & - & 60.42 \\
\midrule
\multirow{3}{*}{\tabincell{c}{\textit{Coding}\\\textit{Tasks}}} & EvalPlus & 69.70 & - & \textbf{73.96} \\
& MultiPL-E & \textbf{65.75} & - & 64.79 \\
& MBPP & 72.60 & - & \textbf{72.60} \\
& CRUX-O & 66.94 & - & \textbf{69.06} \\
\midrule
\multirow{1}{*}{\tabincell{c}{\textit{Multilingual}\\\textit{Tasks}}} & MGSM & 78.75 & - & \textbf{79.93} \\
& INCLUDE & \textbf{65.17} & - & 64.73 \\

\midrule
\tabincell{c}{\textit{College-level}\\\textit{Problems}} & MMMU\textsubscript{val} & - & 57.22 & \textbf{59.33} \\
\midrule
\multirow{1}{*}{\tabincell{c}{\textit{General Visual}\\\textit{Question Answering}}} & MMStar & - & 67.2 & \textbf{69.6} \\
& RealWorldQA\textsubscript{avg} & - & \textbf{73.98} & 71.89 \\
\midrule
\multirow{5}{*}{\tabincell{c}{\textit{OCR-related}\\\textit{Tasks}}} & AI2D & - & 85.88 & \textbf{86.62} \\
& TextVQA\textsubscript{val} & - & \textbf{81.67} & 81.65 \\
& DocVQA\textsubscript{test} & - & 95.19 & \textbf{95.27} \\
& InfoVQA\textsubscript{test} & - & 81.17 & \textbf{83.31} \\
& ChartQA\textsubscript{test Avg} & - & 87.12 & \textbf{87.52} \\
& OCRBench & - & 85.8 & \textbf{86.0} \\
\midrule
\multirow{2}{*}{\tabincell{c}{\textit{Video Understanding}\\\textit{Tasks}}} & Video-MME\textsubscript{w/o sub} & - & 69.22 & \textbf{69.25} \\
& MVBench & - & \textbf{71.87} & 69.50 \\
& LVBench & - & 48.61 & \textbf{51.07} \\
\bottomrule
\end{tabular}
}
\end{table}

%% file: content/conclusion.tex
\section{Conclusion}
\label{sec:conclusion}
In this paper, we introduce Qwen3-Omni-30B-A3B, Qwen3-Omni-30B-A3B-Thinking, Qwen3-Omni-Flash-Instruct, and Qwen3-Omni-Flash-Thinking models. Qwen3-Omni-30B-A3B matches or surpasses the latest same-size unimodal Qwen models on text and vision benchmarks. Notably, on audio processing and dialogue benchmarks, it attains state-of-the-art performance among open-source systems on 32 benchmarks and is comparable to, or better than, the strong proprietary counterpart Gemini-2.5-Pro. The Qwen3-Omni-30B-A3B Thinking variant achieves further gains on complex tasks spanning text, vision, and audio-visual reasoning. Beyond accuracy, the model supports 119 text languages, 19 languages for speech recognition and 10 languages for speech synthesis, and enables audio understanding and interactive sessions up to 40 minutes. Thanks to its streaming architecture and multi-codebook design, Qwen3-Omni at the 30B-A3B scale still delivers an end-to-end first-packet latency of 234 ms.

Research fields often cycle between specialization and integration. In this context, we believe Qwen3-Omni represents a milestone: to our knowledge, it provides the first evidence that fully integrated, end-to-end multimodal training can be achieved without degrading core language capability and other modalities. We are eager to share these findings with the community and hope they will stimulate further research.

For practical usage, Qwen3-Omni-30B-A3B offers strong text and vision capabilities, robust and reliable ASR, interactive speech support in over 20 languages, very low first-packet latency for interactive use, and stable, naturalistic speech synthesis. Crucially, it exhibits advantages over cascaded pipelines, including stronger cross-modal reasoning, lower end-to-end latency, and lower system complexity and cost. In future work, we will further advance the model along multiple axes, including multi-speaker ASR, video OCR, audiovisual proactive learning, and enhanced support for agent-based workflows and function calling.

%% file: content/authors.tex
\section{Authors}
\textbf{Core Contributors:} Jin Xu, Zhifang Guo, Hangrui Hu, Yunfei Chu, Xiong Wang, Jinzheng He, Yuxuan Wang, Xian Shi, Ting He, Xinfa Zhu, Yuanjun Lv, Yongqi Wang, Dake Guo, He Wang, Linhan Ma, Pei Zhang, Xinyu Zhang, Hongkun Hao, Zishan Guo, Baosong Yang, Bin Zhang, Ziyang Ma, Xipin Wei, Shuai Bai, Keqin Chen, Xuejing Liu, Peng Wang, Mingkun Yang, Dayiheng Liu, Xingzhang Ren, Bo Zheng, Rui Men, Fan Zhou, Bowen Yu, Jianxin Yang, Le Yu, Jingren Zhou, Junyang Lin

\textbf{Contributors\footnote{Alphabetical order.}:} An Yang, Anfeng Li, Bei Chen, Beichen Zhang, Bin Lin, Binyuan Hui, Bohan Wang, Buxiao Wu, Chenfei Wu, Cheng Chen, Chen Qiang, Chenhan Yuan, Chenhao Li, Chenxu Lv, Chujie Zheng, Daren Chen, Dayiheng Liu, Dake Guo, Fei Huang, Gezhengyang Zhu, Guangdong Zhou, Hang Zhang, Hongjian Tu, Humen Zhong, Jialong Zuo, Jianhong Tu, Jianwei Zhang, Jiayi Leng, Jing Zhou, Jingren Zhou, Kai Dang, Kexin Yang, Kun Yan, Laiwen Zheng, Lei Xie, Lianghao Deng, Lingchen Meng, Mei Li, Miao Hong, Mingfeng Xue, Minsheng Li, Mingze Li, Peiyang Zhang, Peng Liu, Pengfei Wang, Ruibin Yuan, Rui Hu, Ruiyang Xu, Qidong Huang, Qin Zhu, Que Shen, Shen Li, Shixuan Liu, Sibo Song, Siqi Zhang, Song Chen, Su Hao, Tianyi Tang, Wenbin Ge, Wentao Yao, Wei Ding, Wei Wang, Xiaodong Deng, Xiaotong Chen, Xiao Li, Xian Yang, Xinyao Niu, Xudong Guo, Xin Le, Xuechun Wang, Xutong Jin, Xuancheng Ren, Yang Fan, Yang Liu, Yang Su, Yantao Liu, Yi Wu, Yichang Zhang, Yilei Chen, Yiming Dong, Yinger Zhang, Yizhong Cao, Yuchong Sun, Yuezhang Wang, Yuhao Wang, Yuqiong Liu, Yuanzhi Zhu, Yuxiang Chen, Yuxuan Cai, Yuxuan Liu, Zeyu Cui, Zheng Li, Zhenghao Xing, Zhenru Zhang, Zihan Qiu, ZiYue Jiang, Zhaohai Li, Zhi Li, Zhibo Yang, Zhihai Wang, Zhipeng Zhou

%% file: content/appendix.tex
% 下面的coding和多语言的细节，考虑放到附录。上面的总表其实也已经有所体现
\section{Appendix}
\label{sec:appendix}

\subsection{More Evaluation on Speech and Music Understanding}
\label{sec:audio-eval}

This section reports the performance of the Qwen3-Omni-thinking model on tasks pertaining to ASR/S2TT and Music. 
As shown in Table \ref{tab:audio_asr_thinking} and \ref{tab:audio_music_thinking}, in the domains of ASR/S2TT and Music understanding, the Qwen3-Omni-Thinking model is outperformed by its Instruct counterpart, which indicates that for these predominantly perception-based tasks, the engagement of sophisticated reasoning processes fails to yield performance gains. In fact, it may even introduce a higher propensity for hallucinations.
\input{posttrain_tables/audio_asrs2tt_thinking.tex}

\input{posttrain_tables/audio_music_thinking.tex}

\subsection{Qualitative Results from Qwen3-Omni-30B-A3B-Captioner}
\label{sec:caption-case}

In this section, we illustrate the performance of our finetuned Qwen3-Omni-30B-A3B-Captioner through three representative case studies. The selected scenarios are designed to test the model's proficiency in: (1) analyzing expressive speech, (2) interpreting complex auditory scenes and sound effects, and (3) captioning composite audio that includes speech, music, and ambient sounds. For reproducibility, the original audio samples can be accessed from our public GitHub repository.

% Define the specific blue color from the image
\definecolor{chatmlblue}{RGB}{0, 0, 255}
% gbsn
\begin{CJK*}{UTF8}{gkai}

\begin{tcolorbox}[
    % Box styling
    colback=lightgray,
    colframe=black,
    boxrule=2pt,
    sharp corners,
    width=\textwidth,
    % Title styling
    title={Analysis of Expressive Speech},
    colbacktitle=black,
    coltitle=white,
    fonttitle=\bfseries\sffamily,
    halign title=center,
    % halign=left
]
% Use a monospaced font for the content
\ttfamily
% First user-assistant exchange
The audio clip opens in a studio setting, marked by a faint, persistent electronic hiss and a subtle low-frequency hum, indicating professional but not entirely noise-free recording conditions. The male speaker, whose voice is delivered in a clear, energetic, and highly theatrical manner, begins with an assertive “对” (“Right!”), delivered with a sharp, rising intonation that immediately establishes a performative, self-aware tone. He continues, “这就是我，万人敬仰的太乙真人” (“This is me, Taiyi Zhenren, admired by thousands”), using exaggerated emphasis and a pronounced rising inflection on “万人敬仰” (“admired by thousands”) to convey mock grandeur and comic bravado. His articulation is crisp, and his voice is rich in midrange frequencies, suggesting close-mic’d recording with a touch of digital reverb for dramatic effect.\\

After a brief, deliberate pause, the speaker shifts to a more self-deprecating yet still playful register: “虽然有点婴儿肥，但也掩不住我逼人的帅气” (“Although I’m a bit chubby, it still can’t hide my impressive handsomeness”). Here, the tone becomes lighter and more conversational, with a subtle downward inflection on “婴儿肥” (“chubby”) and a playful, rising emphasis on “逼人的帅气” (“impressive handsomeness”), underscoring the comedic contrast between modesty and self-praise. The delivery is marked by expressive inflection and clear articulation, maintaining a consistent presence in the mix.\\

Throughout the speech, the recording remains clean and undistorted, with no background noises, environmental cues, or audience sounds. The voice is centrally positioned in the stereo field, and the use of reverb and EQ creates a sense of theatricality and spatial depth, while the electronic hiss and hum are subtle but present, hinting at professional studio equipment and post-processing. The clip concludes with a sudden, complete cut to silence, with no trailing sounds or fade-out, indicating a deliberate and precise edit.\\

In summary, this audio features a single male speaker delivering a self-introductory monologue in Mandarin, portraying the legendary Taoist figure Taiyi Zhenren in a comedic, over-the-top manner. The performance blends exaggerated bravado with self-deprecating humor, employing dramatic inflection, expressive pacing, and professional studio production to create a memorable, entertaining character portrayal. The absence of background noise and the clean, intentional editing reinforce the clip’s crafted, theatrical nature, making it suitable for online content, podcasts, or digital entertainment aimed at a modern, culturally literate audience.

\end{tcolorbox}
\end{CJK*}

% Define the specific blue color from the image
\definecolor{chatmlblue}{RGB}{0, 0, 255}
% gbsn
\begin{CJK*}{UTF8}{gkai}

\begin{tcolorbox}[
    % Box styling
    colback=lightgray,
    colframe=black,
    boxrule=2pt,
    sharp corners,
    width=\textwidth,
    % Title styling
    title={Analysis of Complex Scene Sound Effect},
    colbacktitle=black,
    coltitle=white,
    fonttitle=\bfseries\sffamily,
    halign title=center,
    % halign=left,
    % before upper={\par}  % <-- Ensures paragraph mode
]
% Use a monospaced font for the content
\ttfamily
% First user-assistant exchange
The audio clip is a highly produced, cinematic soundscape, lasting 25 seconds, designed to evoke intense drama and imminent danger. It begins with a deep, resonant musical drone and a sharp, metallic clank, immediately establishing a mood of tension and foreboding. A slow, rhythmic, and ominous beat joins, layered with swelling orchestral strings and synthetic textures that build suspense and a sense of scale.\par

Around the 9-second mark, the soundscape explodes with the thunderous, mechanical roar of a massive engine, its deep, percussive pulses dominating the mix. This is accompanied by a high-pitched, metallic screech, suggesting immense machinery in motion or under strain. The orchestral music intensifies, becoming more urgent and climactic, with the addition of a high-frequency electronic whine that adds to the feeling of technological threat.\\

At the 15-second point, a colossal, explosive impact occurs, characterized by a sharp, high-frequency crack followed by a deep, resonant boom. The sound of shattering and debris fills the space, with reverberation suggesting a vast, hard-walled environment—perhaps a hangar, tunnel, or industrial facility. The orchestral music reaches its peak, becoming chaotic and overwhelming, before abruptly ceasing at the 19-second mark.\\

Immediately after the impact, the focus shifts to the sounds of a single person: heavy, strained breathing, the rustle of heavy fabric, and a low groan. These sounds, recorded with close-mic clarity and natural reverberation, indicate the individual is recovering from exertion or injury, likely amidst the aftermath of the catastrophic event. There are no spoken words, but the vocalizations and physical sounds strongly suggest the presence of a male character.\\

Throughout the clip, there are no voices, dialogue, or environmental cues that would suggest a specific culture, language, or geographic setting. The orchestral and electronic musical elements are universal to modern action, sci-fi, and thriller genres, and the production values are consistent with contemporary Hollywood or AAA video game trailers.\\

In summary, the audio is a tightly crafted, non-verbal scene of a catastrophic event—likely a crash, explosion, or mechanical disaster—in a vast, industrial or futuristic setting, immediately followed by the physical and emotional aftermath of a single survivor. The absence of speech, combined with the orchestration, sound design, and spatial cues, positions the clip as a generic but highly effective trailer or scene cue for a dramatic, action-oriented media product.\\

\end{tcolorbox}
\end{CJK*}

% Define the specific blue color from the image
\definecolor{chatmlblue}{RGB}{0, 0, 255}
% gbsn
\begin{CJK*}{UTF8}{gkai}

\begin{tcolorbox}[
    % Box styling
    colback=lightgray,
    colframe=black,
    boxrule=2pt,
    sharp corners,
    width=\textwidth,
    % Title styling
    title={Analysis of Mixed Speech, Audio, and Music},
    colbacktitle=black,
    coltitle=white,
    fonttitle=\bfseries\sffamily,
    halign title=center,
    % halign=left,
    % before upper={\par}  % <-- Ensures paragraph mode
]
% Use a monospaced font for the content
\ttfamily
% First user-assistant exchange
The audio begins with a deep, resonant metallic clang, immediately followed by a powerful, low-frequency rumble that evokes the sound of a massive engine or heavy machinery. The space is acoustically vast, with long reverberation tails and a faint, high-pitched whine, suggesting a colossal, metallic environment such as a spaceship hangar or a vast industrial chamber. A mechanical whirring sound emerges, layered with metallic clanking and grinding, creating the impression of a large vehicle or portal activating. These sounds are punctuated by a sharp, high-frequency crackle, reminiscent of electrical arcs or energy discharges, further reinforcing the technological setting. The mechanical noises intensify, with rapid, high-pitched metallic screeches and impacts that imply the movement or collision of massive metal components.\\

Suddenly, the mechanical sounds fade, and the ambient rumble persists, revealing a subtle, high-frequency hiss—possibly from an air filtration system or atmospheric turbulence—while the vastness of the space remains evident through lingering echoes. A female voice, distant and high-pitched, emerges with a plaintive, childlike tone, asking, "Are we there yet?" Her speech is slightly muffled and reverberant, indicating she is physically separated from the microphone, likely inside the vehicle or machinery. This is followed by a deeper, gravelly male voice, close to the microphone, responding with a gruff, impatient tone: "We get there when we get there." His voice is clear and assertive, contrasting with the female’s, and the exchange is typical of familial banter.\\

The mechanical rumble swells again, joined by a whooshing sound as if air is rushing past, and a rapid metallic clatter signals the rapid movement of machinery or vehicles. The environment is further emphasized by a sharp, high-frequency crackle, suggesting an energy surge or system overload. A third male voice, energetic and friendly, calls out from a moderate distance: "How you doing, honey?" His tone is warm and affectionate, with a slight echo, and the use of "honey" implies a familial relationship. Immediately after, the female voice, now closer and more urgent, responds with a high-pitched, exasperated tone: "Do I have to answer?" Her delivery is quick, sharp, and filled with playful annoyance, reflecting a familiar and comfortable dynamic among the group.\\

As the mechanical sounds subside, a low-frequency hum remains, and the audio transitions into a brief, synthesized musical sting. This consists of a single sustained note from a low-frequency synthesizer, likely a bass or synth pad, which is cut off abruptly, suggesting the end of the scene or a transition to another segment. Throughout, the audio is of high fidelity, with no distortion or noise, and each sound is distinct and well-defined. The spatial characteristics—distance, direction, and reverberation—contribute to a vivid sense of a large, metallic, and technological environment. The dialogue is clear and expressive, with emotional tones ranging from impatience and warmth to playful annoyance. The use of "honey" and the familial banter reinforce the impression of a close-knit group, likely family members, engaged in a shared journey within a science fiction or fantasy context.\\

In summary, the audio presents a dynamic, high-fidelity soundscape of a massive, metallic environment—possibly a spaceship or futuristic vehicle—where a group of family members engage in playful banter as they travel together. Mechanical sounds, spatial cues, and expressive dialogue combine to create a vivid sense of place and character, culminating in a synthesized musical sting that signals a narrative transition. The scene is rich in emotional nuance and technological detail, firmly situating the listener within a science fiction or fantasy setting.

\end{tcolorbox}
\end{CJK*}

%% file: posttrain_tables/audio_asrs2tt_thinking.tex
% \usepackage{booktabs}
% \usepackage{graphicx}
\begin{table}[H]
\centering
\caption{\textbf{Transcription performance for Audio$\to$Text tasks (ASR \& S2TT), comparing Qwen3-Omni-Thinking with the baselines. The highest scores are shown in bold.}}
\label{tab:audio_asr_thinking}
\resizebox{\textwidth}{!}{%
\begin{threeparttable}
\begin{tabular}{@{}lcccccccc@{}}
\toprule
 & \begin{tabular}[c]{@{}c@{}}\textbf{Seed}\\ \textbf{-ASR}\end{tabular} 
& \begin{tabular}[c]{@{}c@{}}\textbf{Voxtral}\\ \textbf{-Mini}\end{tabular} 
& \begin{tabular}[c]{@{}c@{}}\textbf{Voxtral}\\ \textbf{-Small}\end{tabular} 
& \begin{tabular}[c]{@{}c@{}}\textbf{GPT-4o}\\ \textbf{-Transcribe}\end{tabular} 
& \begin{tabular}[c]{@{}c@{}}\textbf{Gemini-2.5}\\ \textbf{-Pro}\end{tabular} 
& \begin{tabular}[c]{@{}c@{}}\textbf{Qwen2.5}\\ \textbf{-Omni}\end{tabular} 
& \begin{tabular}[c]{@{}c@{}}\textbf{Qwen3-Omni}\\ \textbf{-30B-A3B-Thinking}\end{tabular} 
& \begin{tabular}[c]{@{}c@{}}\textbf{Qwen3-Omni}\\ \textbf{-Flash-Thinking}\end{tabular} \\ \midrule
\multicolumn{9}{c}{\textit{EN \& ZH ASR (wer)}} \\ \midrule
\begin{tabular}[c]{@{}l@{}}Wenetspeech\\ \textit{net}|\textit{meeting}\end{tabular} & 4.66 | 5.69 & 24.30|31.53 & 20.33|26.08 & 15.30|32.27 & 14.43|13.47 & 5.91|7.65 & 6.16|8.17 & 6.85|8.42 \\
\begin{tabular}[c]{@{}l@{}}Librispeech\\ \textit{clean} | \textit{other}\end{tabular} & 1.58 | 2.84 & 1.88|4.12 & 1.56|3.30 & 1.39|3.75 & 2.89|3.56 & 1.74|3.45 & 2.22|4.38 & 1.82|4.01 \\
CV15-en & - & 9.47 & 7.79 & 10.01 & 9.89 & 7.61 & 10.44 & 10.52 \\
CV15-zh & - & 24.67 & 19.30 & 9.84 & 8.00 & 5.13 & 6.25 & 6.61 \\
Fleurs-en & 3.40 & 3.96 & 3.77 & 3.32 & 2.94 & 3.77 & 3.75 & 3.67 \\
Fleurs-zh & 2.69 & 12.22 & 7.98 & 2.44 & 2.71 & 2.54 & 2.73 & 2.57 \\ \midrule
\multicolumn{9}{c}{\textit{Multilingual ASR (wer)}} \\ \midrule
\begin{tabular}[c]{@{}l@{}}Fleurs-avg\\ (19 lang)\tnote{a}\end{tabular} & - & 15.67 & 8.09 & 4.48 & 5.55 & 14.04 & 8.63 & 8.88 \\ \midrule
\multicolumn{9}{c}{\textit{Lyric ASR (wer)}} \\ \midrule
MIR-1K (vocal-only)\tnote{b} & 6.45 & 23.33 & 18.73 & 11.87 & 9.85 & 8.15 & 11.15 & 10.47 \\
Opencpop-test & 2.98 & 31.01 & 16.06 & 7.93 & 6.49 & 2.84 & 6.11 & 4.52  \\ \midrule
\multicolumn{9}{c}{\textit{S2TT (BLEU)}} \\ \midrule
Fleurs-en2xx\tnote{c} & - & 30.35 & 37.85 & - & \textbf{39.25} & 29.22 & 36.24 & 36.04  \\
Fleurs-xx2en & - & 27.54 & 32.81 & - & \textbf{35.41} & 28.61 & 30.50 & 30.22 \\
Fleurs-zh2xx & - & 17.03 & 22.05 & - & \textbf{26.63} & 17.97 & 23.74 & 23.77  \\
Fleurs-xx2zh & - & 28.75 & 34.82 & - & \textbf{37.50} & 27.68 & 34.51 & 34.49 \\ \bottomrule\hline
\end{tabular}%
\begin{tablenotes} 
    \item[a] These 19 languages include Arabic, Cantonese, Chinese, Dutch, English, French, German, Indonesian, Italian, Japanese, Korean, Malay, Portuguese, Russian, Spanish, Thai, Turkish, Urdu, Vietnamese.
    \item[b] Transcription is converted into Simplified Chinese.
    \item[c] The results encompass translations across 15 languages: Arabic, Cantonese, Chinese, English, French, German, Indonesian, Italian, Japanese, Korean, Portuguese, Russian, Spanish, Thai, Vietnamese. For notation, ``en2xx'' denotes translation from English into each of the other 14 target languages, where ``xx'' ranges over the remaining language codes.
\end{tablenotes}
\end{threeparttable}
}
\end{table}

%% file: posttrain_tables/audio_music_thinking.tex
% \usepackage{booktabs}
% \usepackage{graphicx}
\begin{table}[H]
\centering
\caption{\textbf{Music understanding performance for Audio$\to$Text tasks, comparing Qwen3-Omni-Thinking with baselines. The highest scores are shown in bold.}}
\label{tab:audio_music_thinking}
\resizebox{\textwidth}{!}{%
\begin{tabular}{@{}lcccccc@{}}
\toprule
 &
  \begin{tabular}[c]{@{}c@{}}\textbf{Best Specialist} \\ \textbf{Models}\end{tabular} &
  \begin{tabular}[c]{@{}c@{}}\textbf{GPT-4o} \\ \textbf{-Audio}\end{tabular} &
  \begin{tabular}[c]{@{}c@{}}\textbf{Gemini-2.5} \\ \textbf{-Pro}\end{tabular} &
  \begin{tabular}[c]{@{}c@{}}\textbf{Qwen2.5} \\ \textbf{-Omni}\end{tabular} &
  \begin{tabular}[c]{@{}c@{}}\textbf{Qwen3-Omni} \\ \textbf{-30B-A3B-Thinking}\end{tabular} &
  \begin{tabular}[c]{@{}c@{}}\textbf{Qwen3-Omni} \\ \textbf{-Flash-Thinking}\end{tabular} \\ \midrule
RUL-MuchoMusic &
  \begin{tabular}[c]{@{}c@{}}47.6 (Audio Flamingo 3)\\ \citep{goel2025audio}\end{tabular} &
  36.1 &
  \textbf{49.4} &
  47.3 &
  48.3 &
  48.4 \\
\begin{tabular}[c]{@{}l@{}}GTZAN\\ \textit{Acc.} \end{tabular} &
  \begin{tabular}[c]{@{}c@{}}87.9 (CLaMP 3)\\ \citep{wu2025clamp}\end{tabular} &
  76.5 &
  81.0 &
  81.7 &
  \textbf{89.0} & \textbf{89.0} \\
\begin{tabular}[c]{@{}l@{}}MTG Genre\\ \textit{Micro F1}\end{tabular} &
  \begin{tabular}[c]{@{}c@{}}\textbf{35.8} (MuQ-MuLan)\\ \citep{zhu2025muq}\end{tabular} &
  25.3 &
  32.6 &
  32.5 &
  32.5 &
  33.0 \\
\begin{tabular}[c]{@{}l@{}}MTG Mood/Theme\\ \textit{Micro F1}\end{tabular} &
  \begin{tabular}[c]{@{}c@{}}10.9 (MuQ-MuLan)\\ \citep{zhu2025muq}\end{tabular} &
  11.3 &
  14.1 &
  8.9 &
  14.3 &
  \textbf{15.4} \\
\begin{tabular}[c]{@{}l@{}}MTG Instrument\\ \textit{Micro F1}\end{tabular} &
  \begin{tabular}[c]{@{}c@{}}\textbf{39.8} (MuQ-MuLan)\\ \citep{zhu2025muq}\end{tabular} &
  34.2 &
  33.0 &
  22.6 &
  36.4 &
  36.4 \\
\begin{tabular}[c]{@{}l@{}}MTG Top50\\ \textit{Micro F1}\end{tabular} &
  \begin{tabular}[c]{@{}c@{}}\textbf{33.2} (MuQ-MuLan)\\ \citep{zhu2025muq}\end{tabular} &
  25.0 &
  26.1 &
  21.6 &
  29.1 &
  29.3 \\
\begin{tabular}[c]{@{}l@{}}MagnaTagATune\\ \textit{Micro F1}\end{tabular} &
  \begin{tabular}[c]{@{}c@{}}\textbf{41.6} (MuQ)\\ \citep{zhu2025muq}\end{tabular} &
  29.2 &
  28.1 &
  30.1 &
  32.2 &
  32.6 \\ \bottomrule
\end{tabular}%
}
\end{table}